\documentclass{article}

\usepackage{graphicx}
\usepackage{subcaption}
\usepackage{bbm}
\usepackage{enumitem}
\usepackage{amsmath,amssymb,amsthm}
\usepackage[export]{adjustbox}
\usepackage{changes}
\usepackage{multirow}

\newcommand{\alert}[1]{\textbf{}}

\newcommand{\BEAS}{\begin{eqnarray*}}
\newcommand{\EEAS}{\end{eqnarray*}}
\newcommand{\BEA}{\begin{eqnarray}}
\newcommand{\EEA}{\end{eqnarray}}
\newcommand{\BEQ}{\begin{equation}}
\newcommand{\EEQ}{\end{equation}}
\newcommand{\BIT}{\begin{itemize}}
\newcommand{\EIT}{\end{itemize}}
\newcommand{\BNUM}{\begin{enumerate}}
\newcommand{\ENUM}{\end{enumerate}}
\newcommand{\BEL}[1]{\begin{equation}\label{#1}}
\newcommand{\EEL}{\end{equation}}

\newcommand{\BA}{\begin{array}}
\newcommand{\EA}{\end{array}}

\usepackage[numbers]{natbib}

    \usepackage[preprint]{neurips_2021}

\usepackage[utf8]{inputenc} 
\usepackage[T1]{fontenc}    
\usepackage{hyperref}       
\usepackage{booktabs}       
\usepackage{amsfonts}       
\usepackage{nicefrac}       
\usepackage{microtype}      
\usepackage{xcolor}         

\usepackage{listings}

\DeclareGraphicsExtensions{.jpg,.png,.PNG,.pdf,.PDF}
\DeclareRobustCommand*{\ora}{\overrightarrow}  

\hypersetup{
 colorlinks=False,
 linkcolor=blue,
 citecolor=blue,
 urlcolor=blue}

\definechangesauthor[color=purple, name=kimin]{KM}

\title{Level 2 Autonomous Driving on a Single Device: Diving into the Devils of Openpilot}

\author{%
  Li Chen$^{1, *, \dagger}$,
  Tutian Tang$^{2, *}$,
  Zhitian Cai$^{1, *}$,
  \textbf{Yang Li}$^{1, *}$,
    \vspace{3pt} \\
\textbf{Penghao Wu}$^{3}$,
  \textbf{Hongyang Li}$^{1,2, \dagger}$,
  \textbf{Jianping Shi}$^{4}$, 
  \textbf{Junchi Yan}$^{1,2}$,
  \textbf{Yu Qiao}$^{1}$
  \vspace{3pt}
  \\
  $^{*}$equal contribution \vspace{.1em} 
  \hspace{4pt}
    $^{\dagger}$project lead
  \vspace{3pt}
  \\
  $^{1}$Shanghai AI Laboratory \vspace{.1em} \hspace{3pt}
  $^{2}$Shanghai Jiao Tong University \vspace{.1em} \hspace{3pt}
  $^{3}$UCSD \vspace{.1em} \hspace{3pt}
  $^{4}$SenseTime
  \vspace{3pt}
  \\
  \texttt{chenli1@pjlab.org.cn} \hspace{4pt}
  \texttt{tttang@sjtu.edu.cn} 
}

\begin{document}

\maketitle

\begin{abstract}

Equipped with a wide span of sensors, 
predominant autonomous driving solutions are becoming more modular-oriented for safe system design. 
Though these sensors have laid a solid foundation, most massive-production solutions up to date still fall into {Level 2 (L2, Partial Driving Automation)} phase. Among these, Comma.ai comes to our sight, claiming one $\$999$ aftermarket device mounted with a single camera and board inside owns the ability to handle L2 scenarios. Together with open-sourced software of the entire system released by Comma.ai, the project is named Openpilot.
%
Is it possible? If so, how is it made possible? With curiosity in mind, we deep-dive into Openpilot
and conclude that its key to success is the end-to-end system design instead of a conventional modular framework.
The model inside Openpilot is briefed as Supercombo, and it can predict the ego vehicle's future trajectory and other road semantics on the fly from monocular input. 
Unfortunately, the training process and massive amount of data to make all these work are \textit{not} available to the public.
To achieve an intensive investigation, we try to reimplement the training details and test the pipeline on public benchmarks. The refactored network proposed in this work is referred to as \textbf{OP-Deepdive}.
For a fair comparison of our version to the original Supercombo,
we introduce a dual-model deployment scheme to test the driving performance
in the real world. 
Experimental results of OP-Deepdive on nuScenes, Comma2k19, CARLA, and in-house realistic scenarios (collected in Shanghai) verify that a low-cost device can indeed achieve most L2 functionalities and be on par with the original Supercombo model.
%
In this report, we would like to share the audience with our latest findings, shed some light on the new perspective of end-to-end autonomous driving from an industrial product-level side, and potentially inspire the community to continue improving the performance based on the environments provided in this work.
Our code, datasets, benchmarks are available at \textcolor{blue}{\url{https://github.com/OpenPerceptionX/Openpilot-Deepdive}}.
\end{abstract}

\clearpage
\tableofcontents
\clearpage

\section{Introduction}\label{sec:intro}
Autonomous driving has witnessed an explosion of solutions, whose capability and capacity never stop growing. On board with the latest technologies in hardware, a single consumer vehicle can feature the ability of performing 176 tera operations per second (teraOPs)~\cite{tops}. It is over 10 billion times more than what the Apollo Guidance Computer can do, the one that took Apollo 11 and astronaut Armstrong to the moon. Such a big step for computational power enables autonomous vehicles to cope with the data generated from various kinds of complex sensors, including multi-view cameras, long-and-short-range radars, LiDARs, precise inertial measurement units, \textit{etc}. A combination of the super-powered computational chip and sensors usually cost more than $10,000$ U.S. dollars.

This appetite for expensive devices successfully drive many self-driving solutions, such as Tesla's AutoPilot~\cite{ingle2016tesla}, Waymo's Waymo One~\cite{lebeau2018waymo}, General Motors's Super Cruise~\cite{motors2013super}, and Baidu's Apollo Go~\cite{fan2018baidu}.
Unfortunately, no matter how exquisitely automotive companies advertise, in the official documents, most massive-production products still belong to \textit{Level 2 (Partial Driving Automation)}~\cite{liu2017computer} driving performance. No one is confident enough to make any further claims beyond L2.

\begin{figure}[t]
    \centering
    \includegraphics[width=\linewidth]{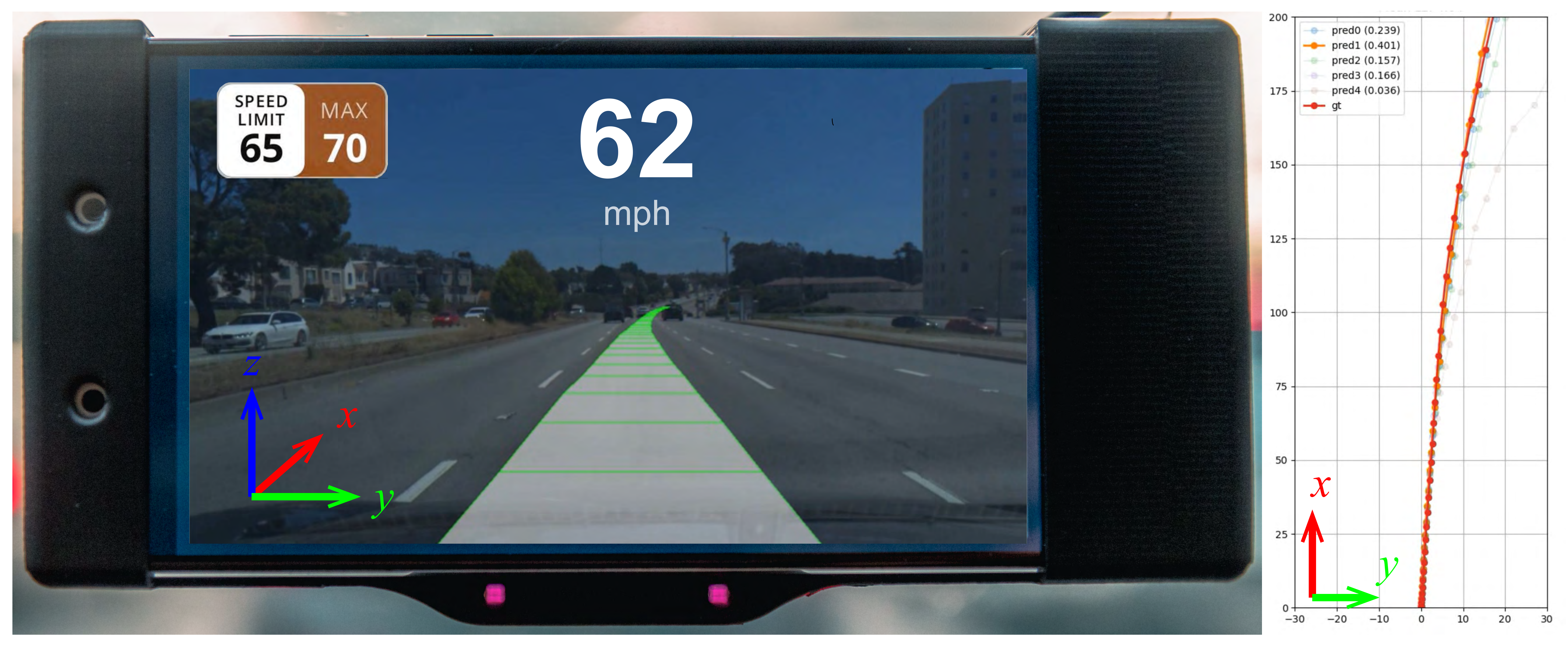}
    \caption{\textbf{Left}: Openpilot achieves L2 with a single device. \textbf{Right}: Predicted trajectories (\texttt{pred0-4}) v.s. human drivers (\texttt{gt}) in bird's-eye view. 
    }
    \label{fig: product}
\end{figure}

However, on the way to commercializing autonomous driving techniques to customers in daily life, Comma.ai\footnote{\url{https://comma.ai/}} builds in their own fashion. Their aftermarket product as depicted in Fig. \ref{fig: product}, Comma Two/Three, is claimed to have the ability of L2, with a single device that costs only $\$999$. It is probably the most affordable device for autonomous driving in the market. Meant to be compatible with many different brands of vehicles, their software, \textbf{Openpilot}, is open-source and hosted on Github~\cite{Openpilot}. More surprisingly, in 2020, Consumer Reports recognized Comma Two as the best among 18 competitors in terms of overall rating\footnote{\url{https://data.consumerreports.org/reports/cr-active-driving-assistance-systems/}}, beating competitors such as Tesla, Cadillac, Ford.

Wait, is it even possible? If so, how is it made possible? What kind of situations does it work in? To answer these questions, we exquisitely test the Comma device, and reimplement part of their systems from scratch. After several months of deepdive, we draw the conclusion that Openpilot is simple and yet effective, for it can work in many simple L2 situations. In this report, we would like to share the audience with our latest findings, and motivate the community to dive into the devils of Openpilot.

\begin{figure}
    \centering
    \includegraphics[width=\linewidth]{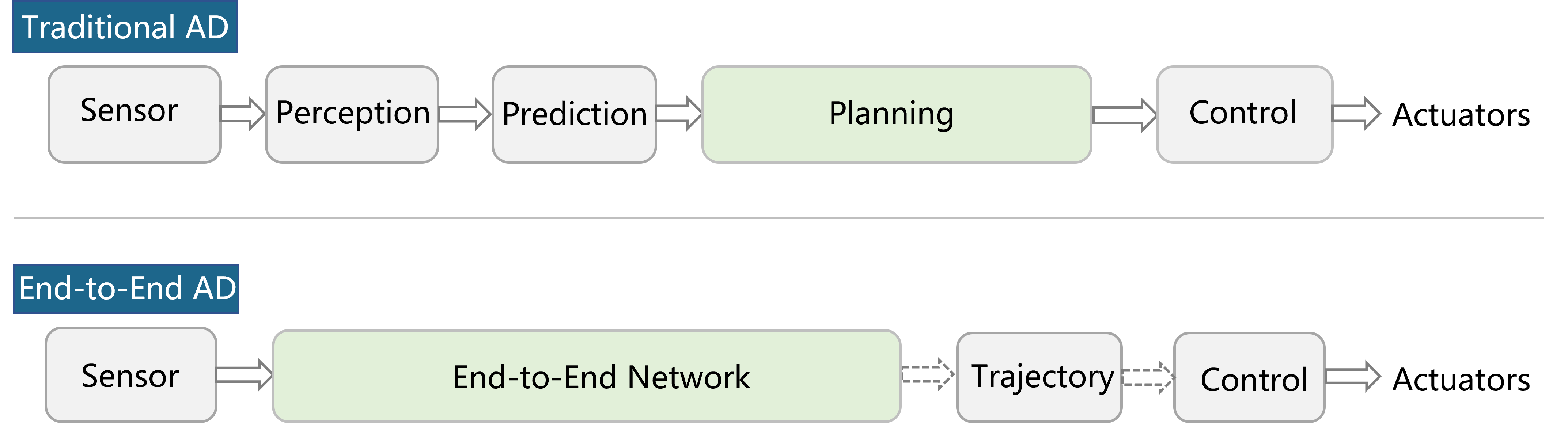}
    \caption{A Comparison of the traditional AD solutions and Openpilot.}
    \label{fig:comparison}
\end{figure}

In contrast to most traditional autonomous driving solutions where the perception, prediction, and planning units are modular-based, Openpilot adopts an end-to-end philosophy on a system-level design to predict the trajectory directly from the camera images. The differences are illustrated in Figure~\ref{fig:comparison}. This end-to-end network in Openpilot is named as \textbf{Supercombo}, and it is the main target that we would like to reimplement and discuss in this work. 

Although Openpilot is deemed as the first to apply the end-to-end spirit for massive consumer products in autonomous driving, there are many pioneering work in academia that try to achieve an end-to-end framework design without explicit perception modules~\cite{casas2021mp3, chitta2021neat, prakash2021transfuser, chen2022lav, hu2022stp3}. These work motivate in sense that end-to-end design would leverage towards better feature representation via co-trained optimization and multi-task learning. Literature in academia achieve impressive results on simulation environments such as CARLA~\cite{dosovitskiy2017carla}, the driving performance and safety guarantee when we transfer algorithms to realistic cases remain unclear, however. 

To this end, the importance of Openpilot lies in the product-level achievement of such an end-to-end design for the whole autonomous driving system, and it can satisfy basic L2 functionalities for daily scenes. The success of Openpilot would encourage more investigation both in academia and industry tryouts.
Unfortunately, the training details as well as massive amount of data are not released to the public. We believe these are very crucial to achieve the L2 goal in an end-to-end spirit.
This motivates us to 
build our own end-to-end model, namely \textbf{OP-Deepdive}, where we keep most of the pre-processes and post-processes consistent with Openpilot. 

Extensive experiments on both public datasets and real-world scenarios are carried out to evaluate the refactored model performance. During the real-world test, we exquisitely analyze the underlying logic and propose a dual-model deployment framework to successfully run our own model on the device.
Numeric results prove that both the original Supercombo and our reimplemented model OP-Deepdive can perform elegant on highway scenarios, but it does not hold true on street scenarios. 
In terms of quantitative metrics, our model is consistent with the original Openpilot model. The on-board performance test further demonstrates that Openpilot is suitable for closed roads (high-speed, ordinary roads, \textit{etc.}), and is excellent for driving assistance system as in ADAS.

In this report, we first give an overview of the Openpilot system, and the test scenarios we use in Section~\ref{sec:pre}. Then, we provide the details of our reimplementation version of OP-Deepdive in Section~\ref{sec:method}. The results, along with the comparison with the original model, are shown in Section~\ref{sec:exp}. Further discussions are noted in Section~\ref{sec:discussion}.
%
We summarize the contributions as follows:
\begin{enumerate}
    \item We test the Openpilot system in real-world scenarios and conclude it can indeed achieve L2 autonomous driving on a single vision device. 

    \item We revivify the training phase of Openpilot from scratch, design the network structure of Supercombo in Openpilot, and test our refactored model on public datasets. 
    OP-Deepdive is comparable with the original one. 
    
    \item This technical report serves as a starting point for the research community to work on the end-to-end system design for the ultimate driving task of planning and control, in an industrial massive to-consumer product sense.
\end{enumerate}

\section{Preliminaries}
\label{sec:pre}

\begin{figure}[t]
    \centering
    \includegraphics[width=0.5\textwidth]{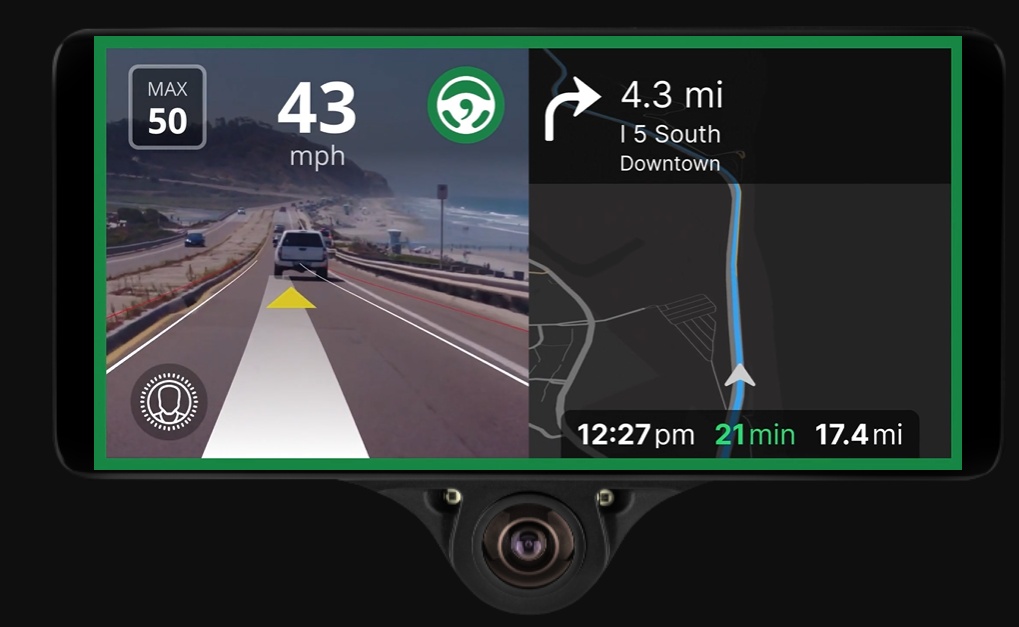}
    \caption{Openpilot on the Comma device. Picture adapted from Comma.ai.}
    \label{fig:op_device}  
\end{figure}

\begin{figure}[t]
    \centering
    \includegraphics[width=\textwidth]{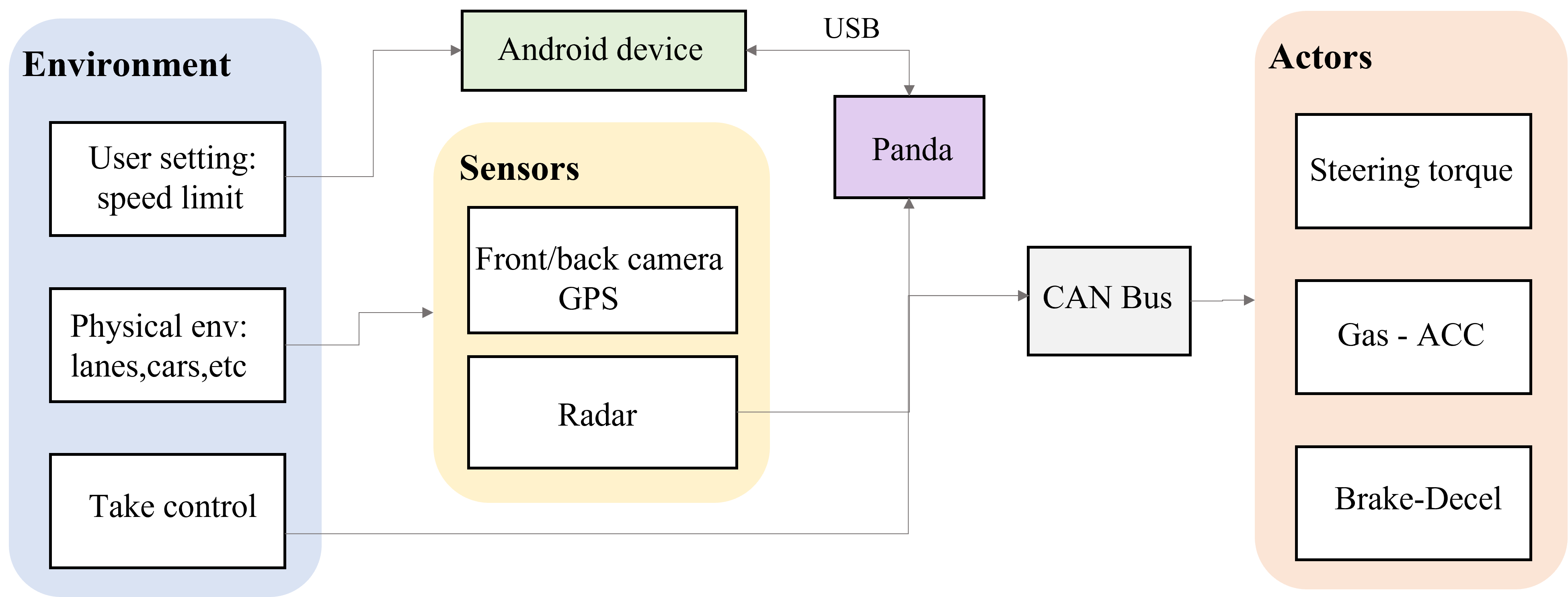}
    \caption{Overview of Openpilot system. Environmental information captured by sensors and states of the ego vehicle exchanged through the Panda device are sent to the customised Android device to run algorithms. Actuator signals are transmitted through CAN Bus to control the vehicle.
    }
    \label{fig:op_overview}
\end{figure}

\subsection{Openpilot: An Overview}

Openpilot is a relatively mature open-source project for L2 assisted driving systems introduced by Comma.ai, which implements conventional assisted driving functions including Adaptive Cruise Control (ACC), Automated Lane Centering (ALC), Forward Collision Warning (FCW) and Lane Keeping Assist (LKA) based on an end-to-end model named Supercombo. The user interface is shown in Figure~\ref{fig:op_device}. It has been made available to massive consumers, and is designed to be compatible with over 150 types of vehicles. Users can enjoy the experience of assisted driving by installing the aftermarket kit on their own car in a few steps.

Openpilot mainly consists of two parts: software and hardware. The software part contains various algorithms and middleware. The hardware device, named EON, acts as the brain of a complete set of devices and is responsible for running the NEO system (a highly-customised Android system) and the software algorithms. The overall system architecture is shown in Figure~\ref{fig:op_overview}.
Different units on the EON device capture the corresponding environmental information, \textit{e.g.}, a camera takes the front image of the physical world and a car interface named Panda extracts the state of the vehicle from CAN Bus. The stock radar on the vehicle is also used in this step.
Then, with these data, Openpilot runs the Supercombo model and validates its output by post-processing software and middleware. It finally sends control signals to the vehicle controller via Panda interface.

We test the system in real world both qualitatively and quantitatively to disclose its strengths and shortcomings. We refer audience for detailed original system-level tests to Appendix.~\ref{sec:appendix} and our webpage \href{https://sites.google.com/view/openpilot-deepdive/home}{\textcolor{blue}{\texttt{https://sites.google.com/view/openpilot-deepdive/home}
}}.

\subsection{The Supercombo Model}

\begin{figure}
    \centering
    \includegraphics[width=\textwidth]{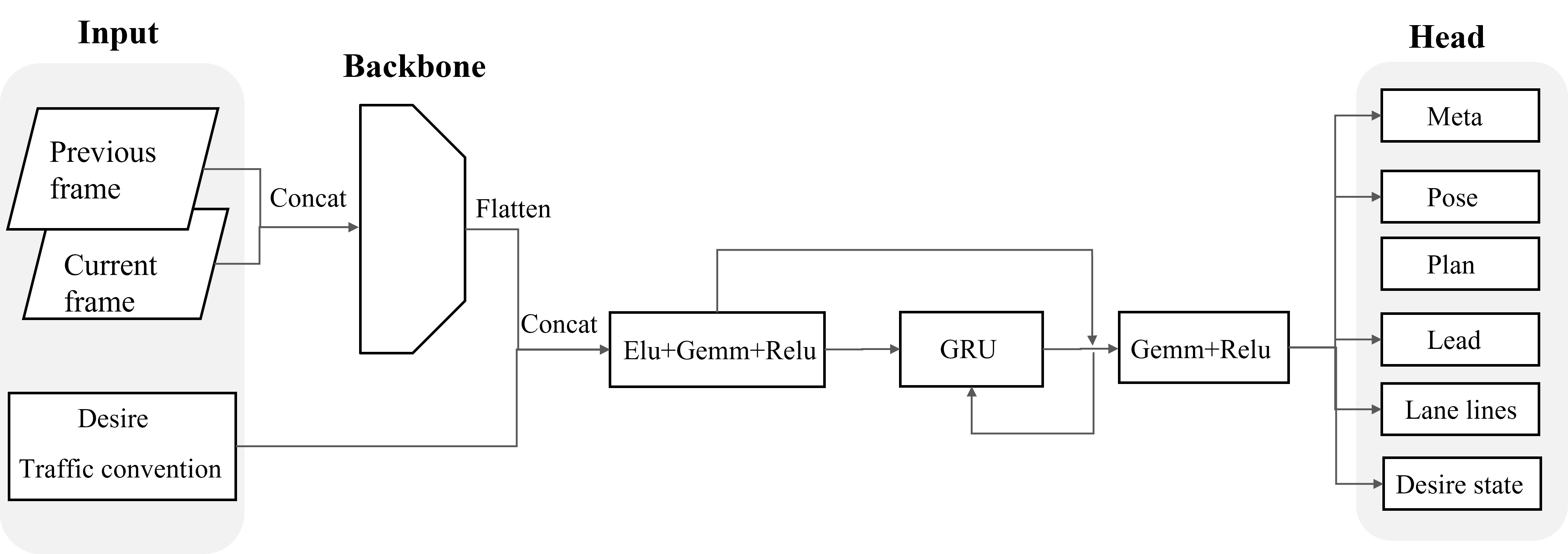}
    \caption{The pipeline of Supercombo. It takes two consecutive frames, a vector representing the desired high level command, and a boolean indicating right/left traffic convention as inputs. Final predictions include a planned trajectory, ego pose, lane lines, state of lead agents, \textit{etc}.}
    \label{fig:Supercombo}
\end{figure}

The core of the whole system is the software algorithms running on the EON device, which are responsible for perception, planning, localization and control. Among these algorithms, the perceptual algorithm (\textit{i.e.}, the Supercombo model) is the key first step that worth digging into. It takes in the images from the camera, and predicts lanes, road edges, location and velocity of leading vehicles, and, most importantly, the trajectory that the ego agent should follow in the future. Hence, it's also called a \textit{trajectory planning model}. We give a simplified illustration of Supercombo in Figure~\ref{fig:Supercombo}. Specifically, the pipeline of the model execution includes the following parts.

\textbf{Preprocessing.} First, the original single-frame 3-channel RGB image taken from the camera with size $(3\times 256\times 512)$, is transformed into 6-channel YUV format with size $(6\times 128\times 256)$. Then, the two consecutive frames is concatenated together as the model input, resulting in a $(12\times 128\times 256)$ input size.

\textbf{Main Network.} The backbone network adopts Google's Efficientnet-B2~\cite{tan2019efficientnet}, which boasts good performance yet high efficiency. It adopts the group convolution to reduce the amount of parameters in the backbone. In order to capture temporal information, a GRU (gated recurrent unit) is connected to the backbone network.

\textbf{Prediction Head.} Several fully connected layers are attached to the GRU, acting as prediction heads. The output includes $5$ possible trajectories, among which the one with the highest confidence is chosen as the planned trajectory. Each trajectory contains the coordinates of $33$ 3D points under ego vehicle's coordinate system. Besides, the Supercombo also predicts lane lines, road edges, the position and velocity of leading objects, and some other information of the vehicle.

To note, although the network structure, along with the pre-and-post processing methods are open-source, the training pipeline and data remain internal. Therefore, we need to reimplement it from scratch. Also, as for the training data, Comma.ai claims that their Supercombo model is trained from a video collection of 1 million minutes of driving\footnote{\url{https://blog.comma.ai/0810release/}}.
However, they only make a small segment of it available to the public.

\section{Method}
\label{sec:method}
In this section, we present an end-to-end trajectory planning model inspired by the Supercombo model, and a deployment framework based on EON device design, as summarized in Figure~\ref{fig:our_model} and Figure~\ref{fig:deployment_framework}.

\subsection{An End-to-end Planning Model}
\begin{figure}
    \centering
    \includegraphics[width=\textwidth]{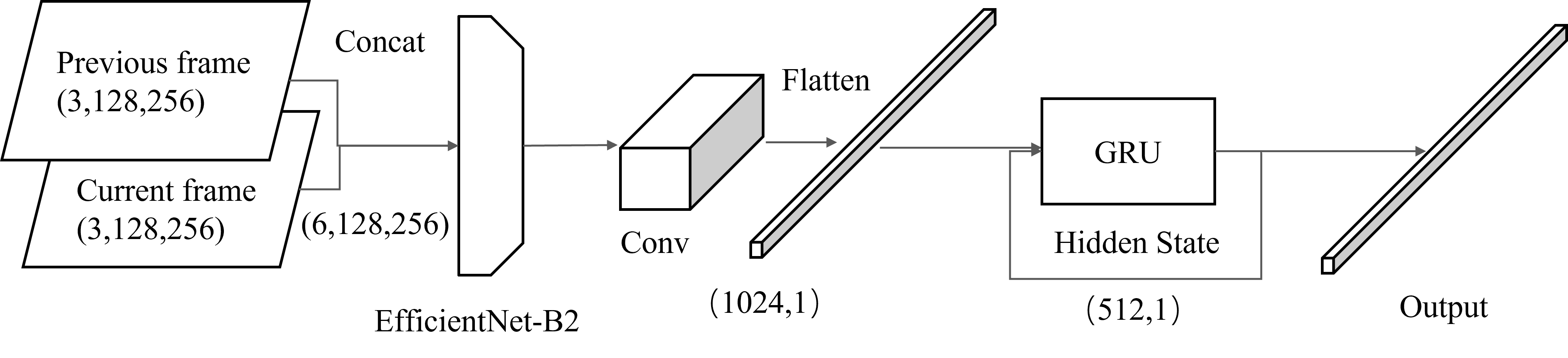}
    \caption{The structure of our reimplemented planning model.}
    \label{fig:our_model}
\end{figure}

Remember our aim is to reimplement the Supercombo model. Ideally, our model should have exactly the same structure as the Supercombo and require the same input and output format. Unfortunately, it is not feasible, for some data required for training is not easily accessible, such as the location and speed of the leading car. Therefore, we make some changes on the input and output format which do not have a significant impact on main conclusions we believe.
As shown in Figure~\ref{fig:our_model}, the main structure of our reimplemented model is quite similar to the original Supercombo. It takes in two consecutive frames from the single, forward-facing camera, and directly predicts the future trajectory. The input frames go through the backbone and are flattened into a feature vector of length 1024. Then, the vector is fed into the GRU module of width 512, enabling the model to remember temporal information. Finally, several fully-connected layers organize the network output into the required format.
More details are discussed below.

\begin{figure}
    \centering
    \includegraphics[width=0.9\textwidth]{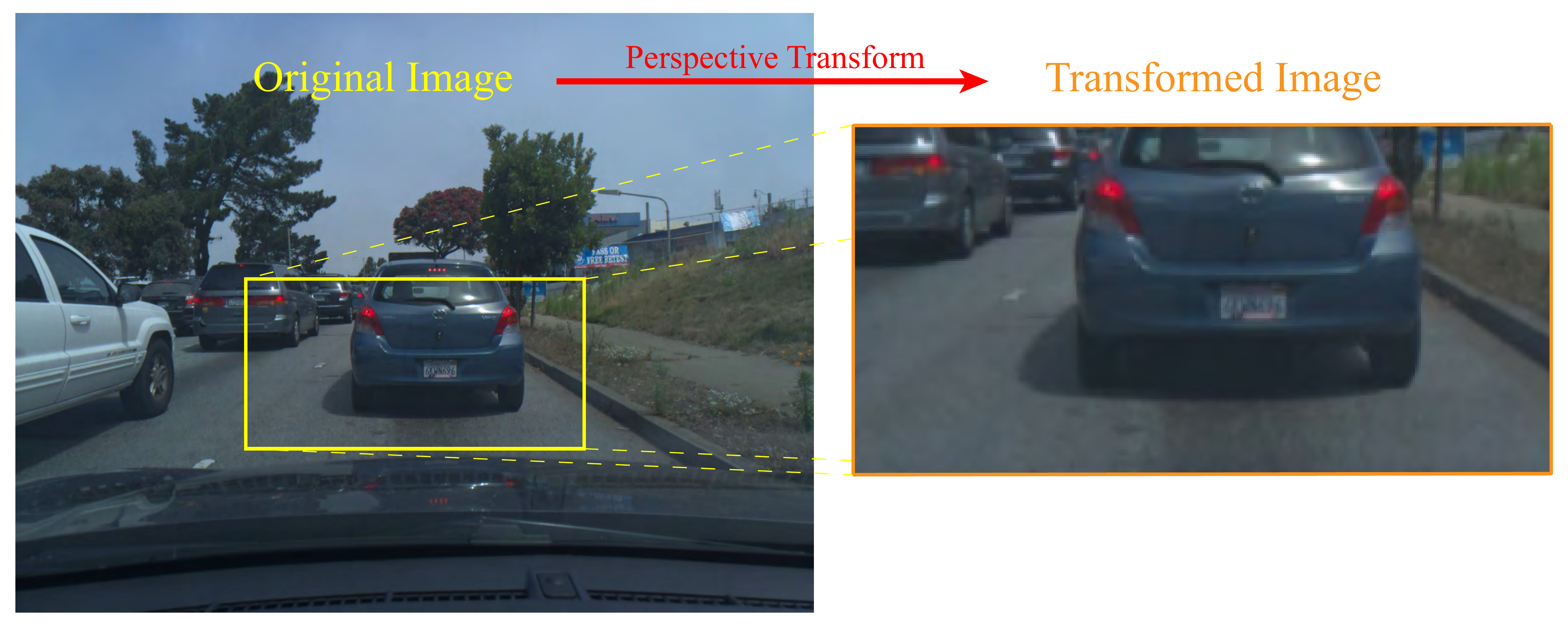}
    \caption{The perspective transformation.}
    \label{fig:perspective_transform}
\end{figure}

\textbf{Preprocessing - Perspective Transformation.}
Openpilot is designed to work with all kinds of vehicles. One direct problem is that different users may mount the device in different positions. This is also called \textit{mount bias}. To eliminate it, it's crucial to know how the cameras are mounted (\textit{i.e.}, the \textit{camera extrinsics}).
Openpilot requires the user to do a camera calibration process after installation, during which the driver should drive the vehicle manually along a long, straight road for a while and keep a relatively fixed speed. The camera extrinsics are derived then. Also, there is an \textit{online calibrating} process that keeps running while the device is working, to keep the extrinsic parameters being updated in case of vibration originated from the suspension system. Once the extrinsics are known, we can do a simple image warping operation, to make the images look like the ones taken from a standard camera. This standard camera is also called the \textit{virtual camera} in some materials. Figure~\ref{fig:perspective_transform} illustrates the result of perspective transformation along with the original image taken from the camera. We can clearly see that Openpilot only pays attention to a very narrow field, for most of the outliers areas in the image are cropped out. The images are resized to $128\times 256$ in this step.

\textbf{Preprocessing - Color Transformation.}
Openpilot directly reads raw images from the CMOS (complementary metal-oxide semiconductor, a kind of image sensors). To convert raw images into the commonly-used RGB-888 format, we need a debayering process, which will introduce extra overhead. Openpilot uses the YUV-422 format as the model input to reduce the overhead. However, as all public-available datasets already recognize RGB-888 as the default image format, it doesn't make sense to convert them back into YUV-422. Therefore, we stick with RGB-888 format. In other words, the color transformation is omitted in our work.

\textbf{Backbone.} Openpilot adopts EfficientNet-B2~\cite{tan2019efficientnet} as the backbone. It features accuracy and efficiency, thanks to the latest AutoML (auto machine learning) and model scaling technologies. It contains 9.2 million parameters, and requires 0.65 billion floating-point operations in one pass. In comparison, the commonly-used ResNet-50~\cite{he2016deep} backbone contains 25 million parameters and requires 2.7 billion floating-point operations in one pass. Given an input tensor of shape $(6, 128, 256)$, the backbone outputs a tensor of shape $(1408, 4, 8)$. Then, a convolutional layer with $3\times 3$ kernel reduces the number of channels to $32$. Finally, the tensor of shape $(32, 4, 8)$ is flattened into a feature vector of length $1024$.

\textbf{GRU Module.} A GRU~\cite{cho2014learning} module with width $512$ is attached following the feature vector. The aim of using GRU here is to capture the temporal information.

\textbf{Prediction Head.} After the GRU module, a few fully-connected layers are responsible to organize the output dimension as required. The Supercombo model will produce a tensor of length 6609 as the final output, which is tremendous. The prediction includes planned trajectories, predicted lane lines, road edges, the position of leading vehicles, and some other information.
However, as currently there is no such dataset that provides all these annotations, we only focus on the planned trajectories in our reimplementation. Let $M$ be the number of possible trajectories. Each trajectory is consisted of the coordinate of $N$ 3D points under ego vehicle coordinate system, and one number for confidence value. Then, the output dimension $D$ is,
\begin{equation}
    D = M\times(N\times3+1).
\end{equation}
Typically, we have $M=5$ and $N=33$.
Considering that the raw value of coordinates under international units is relatively large, we add an exponential function to all $x$ coordinates and add a sinh function to all $y$ coordinates. (Openpilot is never designed to work when reversing!)

\begin{figure}
    \hfill
    \begin{minipage}[c]{0.25\textwidth}
        \includegraphics[width=\textwidth]{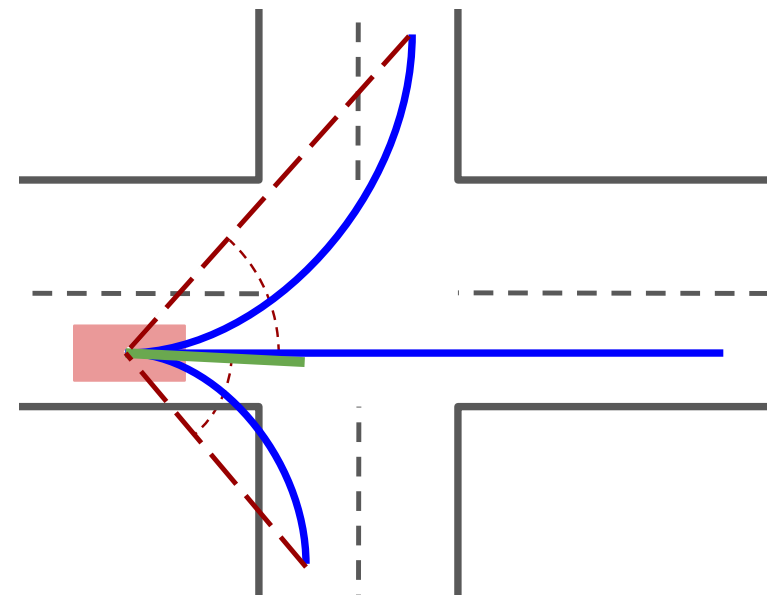}
    \end{minipage}
    \hfill
    \begin{minipage}[c]{0.5\textwidth}
        \caption{How MTP loss works~\cite{mtp_loss}. In this bird's-eye view (BEV), the \textcolor{pink}{pink} rectangle represents the ego vehicle. The ground truth trajectory is shown in \textcolor{teal}{green}. The predicted trajectories are shown in \textcolor{blue}{blue}. It's obvious that we should choose the trajectory that goes straight as $P$.}
    \label{fig:mtp_loss}
    \end{minipage}
\end{figure}

\textbf{Loss.} First, we consider the ground truth trajectory to be the one that human drivers actually drive. Then, we follow the informative Comma.ai's blog
\footnote{\url{https://blog.comma.ai/end-to-end-lateral-planning/}}
to use MTP (Multimodal Trajectory Prediction) loss~\cite{mtp_loss}, which includes a regression loss $\mathcal{L}_{reg}$ and a classification loss $\mathcal{L}_{cls}$. The overall loss is depicted as,
\begin{equation}
    \mathcal{L} = \mathcal{L}_{reg} +\alpha \mathcal{L}_{cls}.
\end{equation}
To be specific, we calculate the cosine similarity between the ground truth and the $M$ predicted trajectories. Denote $P$ as the predicted trajectory with the highest similarity. Then we calculate the regression loss, smooth-$\mathcal{L}_1$, between $P$ and the ground truth. As for $\mathcal{L}_{cls}$, we give $P$ confidence 1, while others are kept 0. The BCE (binary cross entropy) loss is adopted for the classification task. This process is illustrated in Figure~\ref{fig:mtp_loss}. The trajectory with the highest confidence is taken as the prediction result during inference.

\subsection{Deployment on Device}
\begin{figure}
    \centering
    \includegraphics[width=\textwidth]{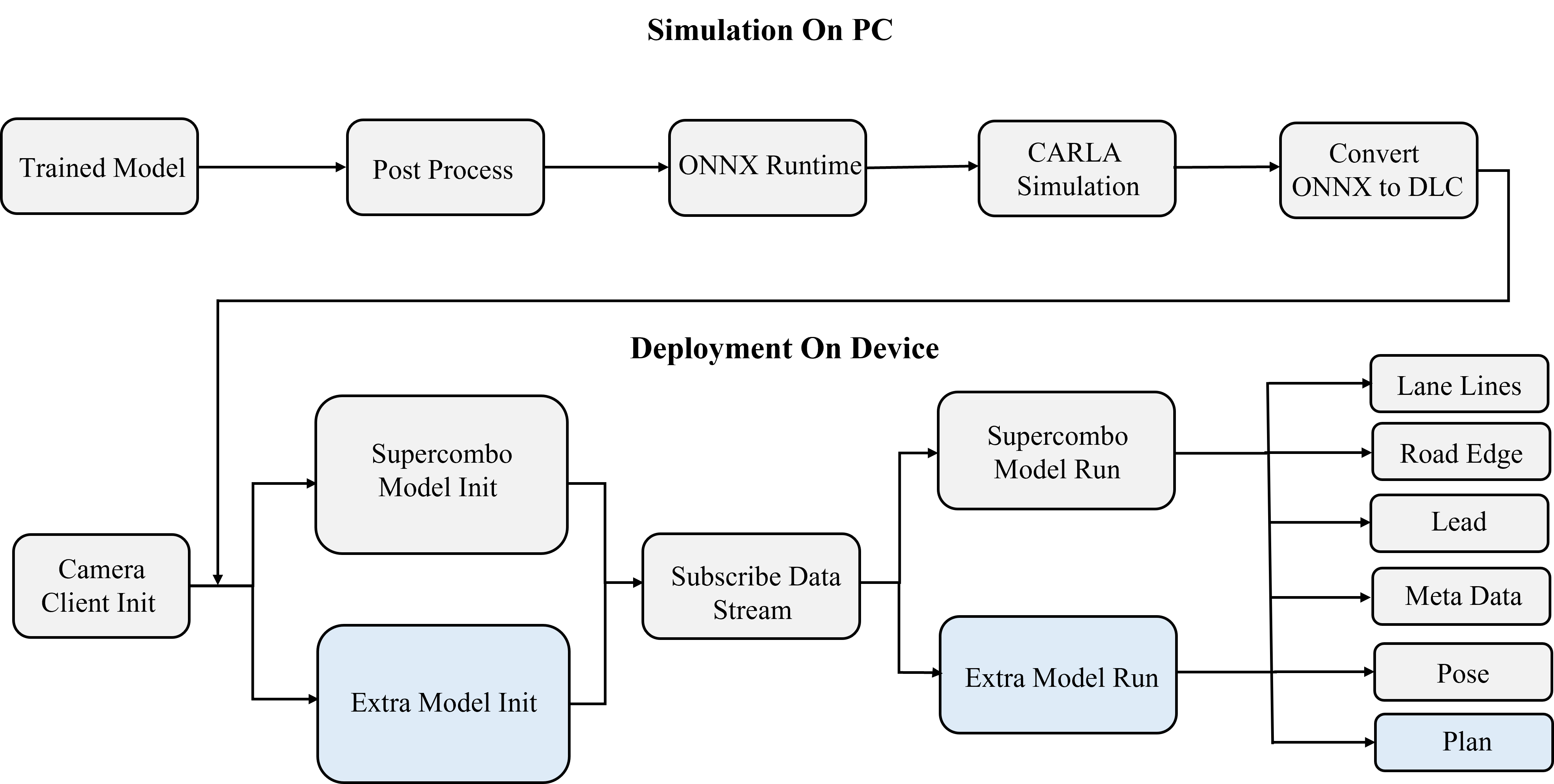}
    \caption{Pipeline of our dual-model deployment framework. Our trained model is validated in CARLA simulation and converted to DLC format before deploying on the device. Two models run simultaneously and each contributes parts of the final outputs. }
    \label{fig:deployment_framework}
\end{figure}

%
To further validate the performance on the real-world application, we explore an on-device deployment pipeline based on the existing Openpilot software and hardware system.
Considering that our model only predicts the trajectory to follow, while the original Supercombo predicts some other information, we design a dual-model deployment scheme. In this scheme, both models are deployed on the board, and make prediction alternately. 
To be specific, we replace the trajectory predicted by the Supercombo with the one predicted by our model, and keep other information untouched.
Since the device-side deployment is tightly coupled touch with the specific hardware platform, we carefully analyze the camera data stream of Openpilot and the rock-bottom model framework in detail. We then add the additional deployed model with the idea of modularity in mind. The overall model deployment framework is shown in Figure~\ref{fig:deployment_framework}. The blue boxes are the part added by our model and replace the plan node in Supercombo.

Besides the design of the overall deployment framework, the following steps in the software development process are essential as well: (a) Apply the model to the Openpilot simulation environment to verify the logic of the dual-model algorithm based on the ONNX model. (b) Convert ONNX to DLC format (the specific model format for neural network acceleration on Qualcomm platform), and migrate the CARLA simulation code to the device side. (c) Connect the Openpilot device to the car and conduct road tests to verify the performance of the dual-model framework. (d) Align the input and output of the reimplemented model with the original one.

\section{Experiments}
\label{sec:exp}

\subsection{Openpilot: A Real-world Test}

In order to test the functions including ACC, ALC, LKA and DM (driver monitoring), we carefully design some test scenarios, study how it actually performs both qualitatively and quantitatively, and try to discover the failure cases.

The main conclusions of our test are as follows. For details, please refer to the Appendix.~\ref{sec:appendix}. 
\begin{enumerate}
    \item Openpilot is capable of running on closed roads (high-speed, ordinary roads, \textit{etc.}), and can perform excellent driving assistance functions.
    
    \item It can not handle some complex urban roads such as two-way roads without lanes. In these cases, human intervention is required to avoid collision.
\end{enumerate}

\subsection{OP-Deepdive: Our Reimplementation of Supercombo}

\subsubsection{Dataset} \label{sec:dataset}
We train and evaluate our model on two datasets, nuScenes~\cite{caesar2020nuscenes} and Comma2k19~\cite{comma2k19}. 
Table~\ref{tab:two_dataset_comp} shows some key features of them.

\begin{table*}[h]
\centering
\caption{Comparison between the two datasets used in this report.
}
\label{tab:two_dataset_comp}
\small
\begin{tabular}{lcccccc}
\toprule
Dataset & \begin{tabular}[c]{@{}c@{}}Raw\\ FPS (Hz)\end{tabular} & \begin{tabular}[c]{@{}c@{}}Aligned\\ FPS (Hz)\end{tabular} & \begin{tabular}[c]{@{}c@{}}Length Per\\ Sequence\\ (Frames/Second)\end{tabular} & \begin{tabular}[c]{@{}c@{}}Altogether\\ Length\\ (Minutes)\end{tabular} & Scenario & Locations \\
\midrule
nuScenes \cite{caesar2020nuscenes} & 12 & 2 & 40 / 20 & 330 & Street & \begin{tabular}[c]{@{}c@{}}America\\ Singapore\end{tabular} \\
\midrule
Comma2k19 \cite{comma2k19} & 20 & 20 & 1000 / 60 & 2000 & Highway & America \\
\bottomrule
\end{tabular}
\end{table*}

\textbf{nuScenes.}
nuScenes~\cite{caesar2020nuscenes} is a public large-scale dataset for autonomous driving curated by Motional. It is multipurpose, supporting various kinds of tasks, like detection, tracking, prediction, and segmentation. It provides 12 raw frames per second, but, only 2 of them are aligned and can be used in our setting. We extract $556$ minutes of valid video clips from it, and randomly split it into $80\%/20\%$ as the training set and validation set, respectively.

To cater for the relatively low $2$ Hz frame rate, we make a change on the ground truth of the trajectory. Instead of $33$, each ground truth trajectory only contains $10$ points that record ego vehicle's position in $5$ seconds in the future.

\textbf{Comma2k19.}
Comma2k19~\cite{comma2k19} is a dataset of over 33 hours of commute in California's 280 highway, presented by Comma.ai in 2018. This dataset was collected using their early products. We believe it's a small subset of the training data on which Comma.ai train Supercombo, which is 1 million minutes as stated in their blog. Therefore, this dataset is ideal for training our model on. Again, we pick $80\%$ as the training set, and the rest $20\%$ as the validation set.

We align the ground truth with Openpilot for Comma2k19. Each ground truth trajectory contains $33$ planned positions in $10$ seconds in the future, under current ego vehicle's coordinate system. The points are not uniformly distributed. Instead, they follow a mechanism called \textit{time-based anchor}.

At an indefinite time $t=0$, the trajectory contains the locations that the car would appear in at some time $T$. $T$ is exactly the time-based anchor. The values for $T$ are fixed, chosen from the set
\texttt{\{0., 0.00976562, 0.0390625, 0.08789062, 0.15625, 
0.24414062, 0.3515625, 0.47851562, 0.625, 0.79101562, 
0.9765625, 1.18164062, 1.40625, 1.65039062, 1.9140625, 
2.19726562, 2.5, 2.82226562, 3.1640625, 3.52539062, 
3.90625, 4.30664062, 4.7265625, 5.16601562, 5.625, 
6.10351562, 6.6015625, 7.11914062, 7.65625, 8.21289062, 
8.7890625, 9.38476562, 10.\}}. Note that they are not uniformly distributed between 0 and 10. Instead, they are dense in the near future and become sparse as the time goes, which suggests that the model should focus more on the near future. The coordinates are linearly interpolated from the 30 Hz data.

\subsubsection{Metrics} \label{sec:metric}

We use two kinds of metrics to evaluate the methods, namely the imitation metrics and the comfort metrics.

\textbf{Imitation Metrics.}
Imitation metrics are aimed to show how well a model learns from the human drivers.
Inspired by the 3D detection metric used in KITTI detection challenge~\cite{Geiger2012CVPR}, we first divide the trajectory points to 5 groups according to the distances on $x$ axis, that is, $0\sim10$, $10\sim20$, $20\sim30$, $30\sim50$ and $50+$ meters. In each range, we calculate two metrics, the average Euclidean distance and average precision.

\begin{itemize}
    \item \textbf{Average Euclidean Distance Error.} As the name suggests, we calculate the Euclidean distances in 3D space between the corresponding predicted and ground truth points. Then, we average them. Besides, we also calculate the Euclidean distances when the points are projected onto $x$ and $y$ axis. The lower, the better. The unit is \textit{meter}.
    
    \item \textbf{Average Precision.} If the distance between two corresponding points is smaller than a threshold, we mark it "Hit". Otherwise, we mark it "Miss". Then, we can calculate the hit rate, and average them. For example, AP@0.5 means the average hit rate under threshold 0.5. The higher the better.
\end{itemize}

\textbf{Comfort Metrics.}
Comfort metrics can reflect if a possible trajectory will make passengers feel comfortable. To be specific, we use jerk and lateral acceleration to measure comfort.

\begin{itemize}

\item \textbf{Jerk.} Jerk is the rate at which an object's acceleration changes with respect to time. To note, since jerk is a directed quantity, we actually report the amplitude of it. The unit is $m/s^3$. The lower, the better.

\item \textbf{Lateral Acceleration.} Lateral acceleration is the acceleration which is transverse to the direction of travel of a car. Also, it's a directed quantity, so we actually report its amplitude. The unit is $m/s^2$. The lower, the better.
\end{itemize}

\subsubsection{Implementation Details}

Our implementation is compatible with PyTorch~\cite{paszke2019pytorch} 1.8. 
By default, we use AdamW as the optimizer. The batch size and the learning rate are set to be 48 and $10^{-4}$, respectively. A gradient clip of value $1.0$ is applied. During training, we use 8 NVIDIA V100 GPUs. Since there is a GRU module, we need to initialize its hidden state by filling zeros. The parameters are updated once every 40 steps, over which the gradients are accumulated. It takes approximate 120 hours to train 100 epochs on Comma2k19 dataset. On a single NVIDIA GTX 1080 GPU, the network can inference at a speed of 100 FPS.

\subsubsection{Quantitative Results} \label{sec:result}

\textbf{On nuScenes Dataset.}
As each ground truth trajectory only lasts 5 seconds and most of the collection scenarios are in busy downtown, there are only a few samples that fall in the ``50+ meter'' range. Therefore, we do not report the results in that range.

Table~\ref{tab:nuscenes_ours} shows the results of our reimplemented model on nuScenes dataset. To be honest, we fail to achieve a reasonable result, for the distance errors are large while the APs are low.

Trying to understand the magic behind, we test the original Supercombo model on nuScenes dataset. As shown in Table~\ref{tab:nuscenes_op}, the results are even worse. This is understandable, for the framerate in nuScenes is not 30 FPS. We further finetune the Supercombo model using our pipeline. As shown in Table~\ref{tab:nuscenes_op_ft}, it performs a bit better than our reimplementation. We conclude that on nuScenes dataset, better pre-trained weights can lead to a better performance, but the benefit is not significant.

Since it's unreasonable to calculate the lateral acceleration and jerk based on such a low frame rate (2 Hz), we do not report comfort metrics on nuScenes dataset.

\begin{table}[t]
\centering
\caption{Results on nuScenes dataset.}
\begin{subtable}{\textwidth}
\centering
\caption{Imitation metrics for our reimplemented model, OP-Deepdive. \textbf{D.E.:} Euclidean Distance Error. \textbf{D.E.}$\ora{x}$: Euclidean distance error projected onto $x$ axis. \textbf{AP@0.5:} Average precision under threshold 0.5 meter. The rest can be inferred.}
\label{tab:nuscenes_ours}
\begin{tabular}{c|cccccc}
\toprule
\begin{tabular}[c]{@{}c@{}}Range\\ (Meter)\end{tabular} & \begin{tabular}[c]{@{}c@{}}D.E. \\ (Meter)\end{tabular} & \begin{tabular}[c]{@{}c@{}}D.E. $\overrightarrow{x}$\\ (Meter)\end{tabular} & \begin{tabular}[c]{@{}c@{}}D.E. $\overrightarrow{y}$\\ (Meter)\end{tabular} & AP@0.5 & AP@1 & AP@2 \\
\midrule
0-10 & 2.02 & 1.92 & \textbf{0.25} & 0.28 & 0.51 & \textbf{0.73} \\
10-20 & 4.39 & 3.91 & 1.01 & 0.045 & 0.14 & 0.33 \\
20-30 & 5.25 & 4.63 & 1.24 & 0.016 & 0.067 & 0.18 \\
30-50 & 6.51 & 5.93 & 1.26 & 0.010 & 0.038 & 0.12 \\
\bottomrule
\end{tabular}
\end{subtable}

\begin{subtable}{\textwidth}
\centering
\caption{Imitation metrics for the original Supercombo model.}
\label{tab:nuscenes_op}
\begin{tabular}{c|cccccc}
\toprule
\begin{tabular}[c]{@{}c@{}}Range\\ (Meter)\end{tabular} & \begin{tabular}[c]{@{}c@{}}D.E. \\ (Meter)\end{tabular} & \begin{tabular}[c]{@{}c@{}}D.E. $\overrightarrow{x}$\\ (Meter)\end{tabular} & \begin{tabular}[c]{@{}c@{}}D.E. $\overrightarrow{y}$\\ (Meter)\end{tabular} & AP@0.5 & AP@1 & AP@2 \\
\midrule
0-10 & 1.96 & 1.85 & \textbf{0.32} & 0.237 & 0.408 & \textbf{0.692} \\
10-20 & 5.80 & 5.42 & 1.04 & 0.025 & 0.064 & 0.170 \\
20-30 & 8.54 & 8.11 & 1.26 & 0.013 & 0.038 & 0.090 \\
30-50 & 11.36 & 11.05 & 1.23 & 0.008 & 0.023 & 0.053 \\
\bottomrule
\end{tabular}
\end{subtable}

\begin{subtable}{\textwidth}
\centering
\caption{Imitation metrics for the finetuned Supercombo model.}
\label{tab:nuscenes_op_ft}
\begin{tabular}{c|cccccc}
\toprule
\begin{tabular}[c]{@{}c@{}}Range\\ (Meter)\end{tabular} & \begin{tabular}[c]{@{}c@{}}D.E. \\ (Meter)\end{tabular} & \begin{tabular}[c]{@{}c@{}}D.E. $\overrightarrow{x}$\\ (Meter)\end{tabular} & \begin{tabular}[c]{@{}c@{}}D.E. $\overrightarrow{y}$\\ (Meter)\end{tabular} & AP@0.5 & AP@1 & AP@2 \\
\midrule
0-10 & 1.39 & 1.31 & \textbf{0.22} & 0.305 & 0.568 & \textbf{0.809} \\
10-20 & 3.57 & 3.16 & 0.84 & 0.054 & 0.162 & 0.387 \\
20-30 & 4.78 & 4.28 & 1.06 & 0.028 & 0.088 & 0.235 \\
30-50 & 6.25 & 5.90 & 0.93 & 0.015 & 0.050 & 0.149\\
\bottomrule
\end{tabular}
\end{subtable}

\end{table}

\textbf{On Comma2k19 Dataset.}
In terms of the imitation metrics, our model achieves very good results on Comma2k19 dataset. The original Supercombo model shows comparable results, as we can see in Table~\ref{tab:comma_ours} and Table~\ref{tab:comma_op}. On average, our model performs better than the original Supercombo.

As for the comfort metrics, however, things become different. As Table~\ref{tab:comfort} shows, the trajectories produced by our model feature higher jerks and lateral accelerations than the ones produced by the Supercombo. In the meantime, the trajectories from the human demo (\textit{i.e.}, the ground truth) show much lower jerks and lateral accelerations than both of the models. This suggests that the predicted trajectories are not smooth enough, compared with human drivers. However, this does not mean that Openpilot is not so good as a human driver, for the predicted trajectories require further processing before Openpilot sends commands to the car.

\begin{table}[t]
\centering
\caption{Results on Comma2k19 dataset.}
\begin{subtable}{\textwidth}
\centering
\caption{Imitation metrics for our reimplemented model, \textbf{OP-Deepdive}.}
\label{tab:comma_ours}
\begin{tabular}{c|cccccc}
\toprule
\begin{tabular}[c]{@{}c@{}}Range\\ (Meter)\end{tabular} & \begin{tabular}[c]{@{}c@{}}D.E. \\ (Meter)\end{tabular} & \begin{tabular}[c]{@{}c@{}}D.E. $\overrightarrow{x}$\\ (Meter)\end{tabular} & \begin{tabular}[c]{@{}c@{}}D.E. $\overrightarrow{y}$\\ (Meter)\end{tabular} & AP@0.5 & AP@1 & AP@2 \\
\midrule
0-10 & 0.451 & 0.426 & \textbf{0.067} & 0.909 & 0.951 & \textbf{0.971} \\
10-20 & 1.161 & 1.093 & 0.180 & 0.589 & 0.808 & 0.914 \\
20-30 & 1.624 & 1.529 & 0.248 & 0.384 & 0.651 & 0.850 \\
30-50 & 2.335 & 2.192 & 0.367 & 0.225 & 0.465 & 0.729 \\
50+ & 7.373 & 6.475 & 2.04 & 0.030 & 0.100 & 0.239 \\
\bottomrule
\end{tabular}
\end{subtable}

\begin{subtable}{\textwidth}
\centering
\caption{Imitation metrics for the original model, \textbf{Supercombo}.}
\label{tab:comma_op}
\begin{tabular}{c|cccccc}
\toprule
\begin{tabular}[c]{@{}c@{}}Range\\ (Meter)\end{tabular} & \begin{tabular}[c]{@{}c@{}}D.E. \\ (Meter)\end{tabular} & \begin{tabular}[c]{@{}c@{}}D.E. $\overrightarrow{x}$\\ (Meter)\end{tabular} & \begin{tabular}[c]{@{}c@{}}D.E. $\overrightarrow{y}$\\ (Meter)\end{tabular} & AP@0.5 & AP@1 & AP@2 \\
\midrule
0-10 & 0.4396 & 0.2679 & \textbf{0.1998} & 0.7966 & 0.9510 & \textbf{0.9829} \\
10-20 & 1.4824 & 1.2157 & 0.2928 & 0.1782 & 0.6170 & 0.8617 \\
20-30 & 2.2831 & 1.8652 & 0.4483 & 0.0263 & 0.2661 & 0.7368 \\
30-50 & 3.4265 & 2.7790 & 0.6818 & 0.0026 & 0.0889 & 0.4922 \\
50+ & 11.1124 & 9.0153 & 2.4796 & 0.0001 & 0.0004 & 0.0062 \\
\bottomrule
\end{tabular}
\end{subtable}

\begin{subtable}{\textwidth}
\centering
\caption{Comfort metrics.}
\label{tab:comfort}
\begin{tabular}{lcccc}
\toprule
Method & Average Jerk & Max Jerk & Average Lateral Acc. & Max Lateral Acc. \\ \midrule
Supercombo & 2.2243 & 10.8209 & 0.3750 & 1.0505 \\
OP-Deepdive (ours) & 4.7959 & 24.007 & 0.4342 & 1.7931 \\
Human Demo & 0.3232 & 2.2764 & 0.3723 & 0.7592 \\
\bottomrule
\end{tabular}
\end{subtable}
\end{table}

\subsection{Dual-model Experiments}

After we successfully reimplement the Supercombo model which achieves a reasonable performance on some public datasets, we would like to deploy it on the Comma Two board and see its real-world performance.

\subsubsection{CARLA Simulation Test}
We first test the dual-model framework in CARLA~\cite{dosovitskiy2017carla}, an autonomous driving simulation platform. The main purpose of the simulation test is to verify whether the dual-model framework is logically correct before we perform a real on-board test. However, please kindly note that there are some gap between simulation and the real world.

\begin{itemize}
    \item The source of data is different. The data stream in CARLA is rendered by the simulator. There is a certain domain gap between the rendered images and the real-world images.
    
    \item The hardware platform is different. The simulation is conducted on the PC, which has sufficient computing power. However, in the real-world test, the Comma Two board is based on the Qualcomm SOC platform, which has limited computing power and requires explicit neural network acceleration.
    
    \item The model format is different. The simulation platform adopts ONNX, which is a commonly-used model format, while the the Comma Two board requires DLC, a specific model format for Qualcomm mobile platforms.
\end{itemize}

\begin{figure}[t]
\centering
\begin{subfigure}{0.57\textwidth}
    \centering
    \includegraphics[width=\textwidth]{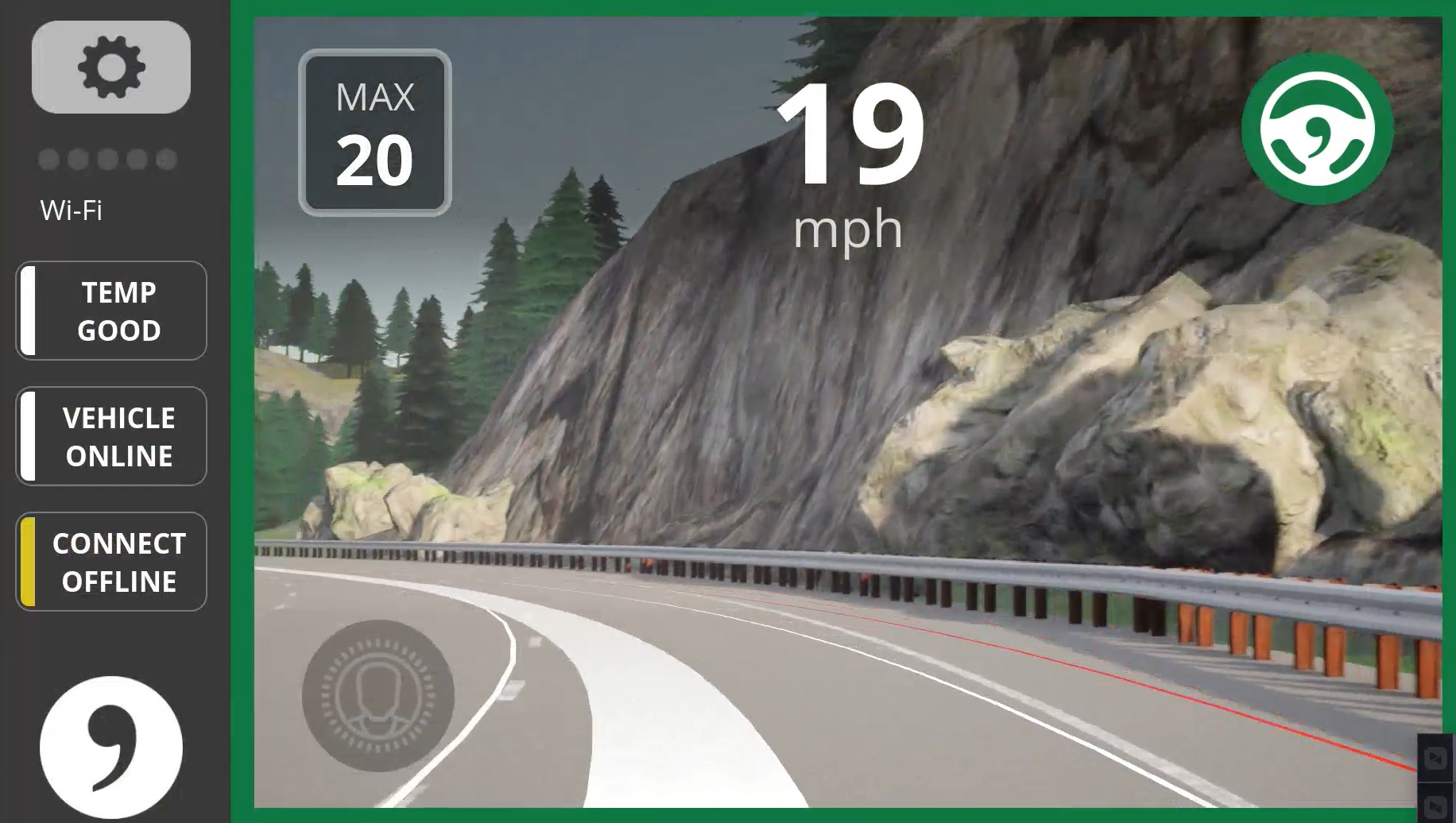}
    \caption{Our dual-model deployment framework running in CARLA.}
    \label{fig:carla_test}
\end{subfigure}
\hfill
\begin{subfigure}{0.4\textwidth}
    \centering
    \includegraphics[width=\textwidth]{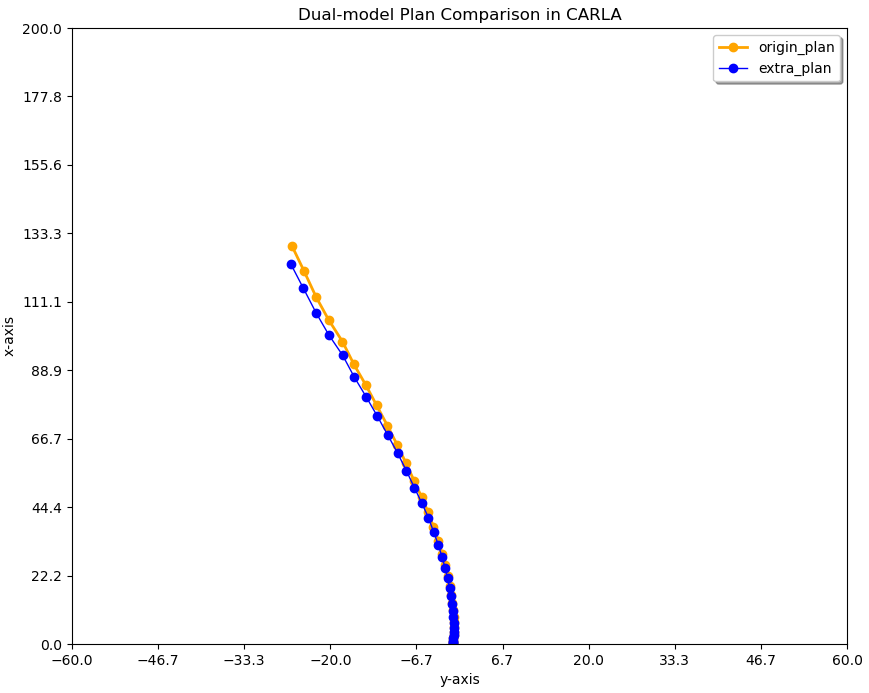}
    \caption{A comparison of trajectories.}
    \label{fig:best_plan_cmp_sim}
\end{subfigure}
\caption{Qualitative results of daul-model deployment framework validated in CARLA.}
\end{figure}

Figure~\ref{fig:carla_test} shows an example of the dual-model framework running in CARLA simulation. Note that the planned trajectory is from our reimplemented model while the lane lines are the original Supercombo model. The two models are running simultaneously.

We further compare the two model's predicted trajectories in Figure~\ref{fig:best_plan_cmp_sim}. Overall, the trajectories are close to each other. However, compared with the original Openpilot, the execution time and frame loss rate increase apparently. The reasons may include two aspects. Two models run simultaneously so that they occupy more hardware resources. In addition, they subscribe to the same data flow node, leading to the increase of the frame loss rate. It is reasonable to suspect that the problem may be further magnified when it is deployed on a mobile device.

\subsubsection{Real-world Test}

\begin{figure}[t]
\centering 
\begin{subfigure}{0.48\textwidth}
    \centering
    \includegraphics[width=\textwidth]{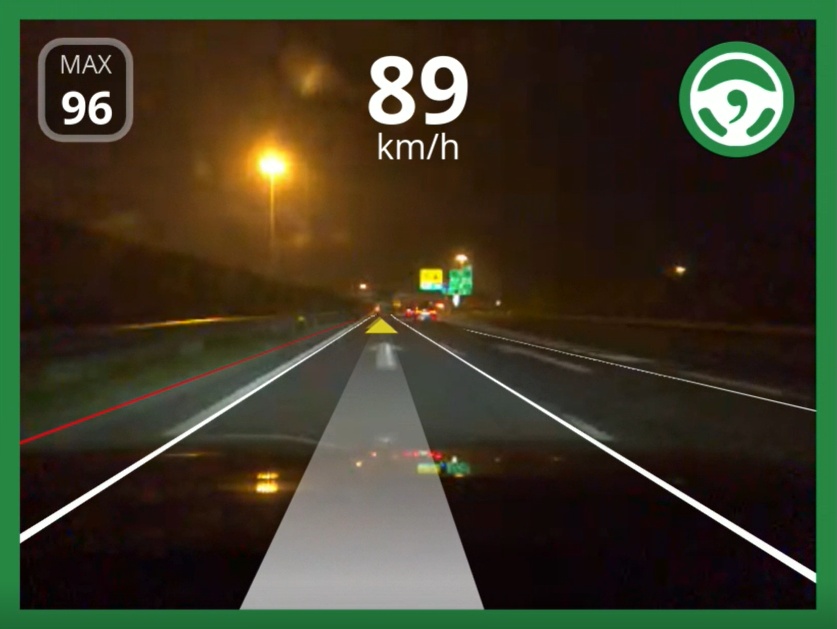}
    \caption{Our dual-model deployment framework on-board test.}
    \label{fig:dual_model_device_test}
\end{subfigure}
\hfill
\begin{subfigure}{0.46\textwidth}
    \centering
    \includegraphics[width=\textwidth]{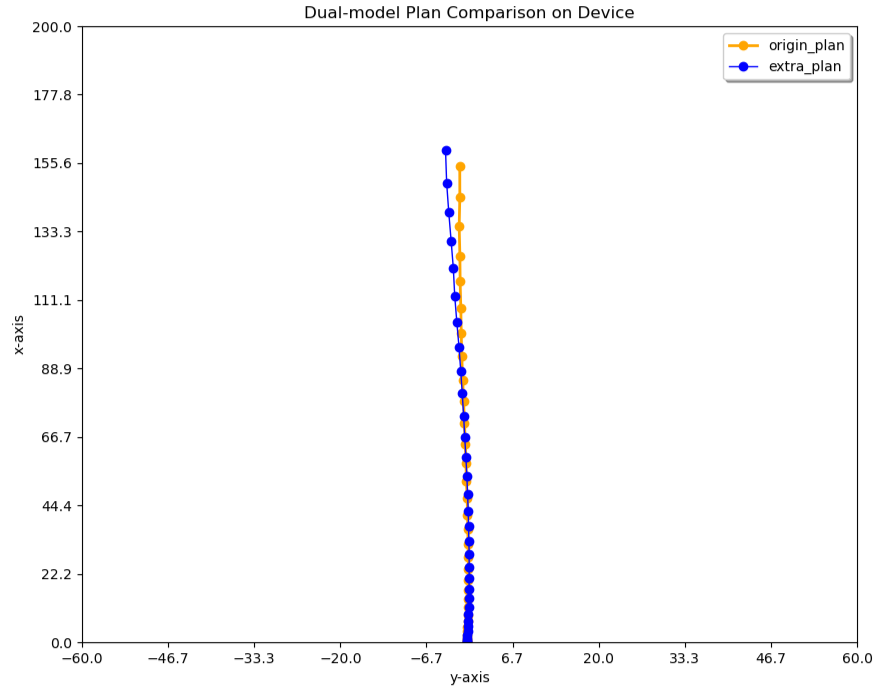}
    \caption{A comparison of trajectories.}
    \label{fig:best_plan_cmp_device}
\end{subfigure}
\caption{Qualitative results of daul-model deployment framework validated onboard.}
\end{figure}

After we verify the dual-model framework in simulation environment, we can finally deploy it on board. The board we use is Comma Two, and the vehicle is Lexus ES 2020.
Figure~\ref{fig:dual_model_device_test} shows the result of our dual-model deployment framework. For safety concern, the software is running with the dubug mode on, where it makes predictions but does not control the vehicle.
Overall, it is in line with expectations, which proves that our deployment scheme on a single device is effective. Similar to the previous section, we also show both trajectories in Figure~\ref{fig:best_plan_cmp_device}. It can be observed that the output curves are relatively consistent.
We further analyze the running state of the two models on device. The running time and frame loss rate are higher than those in the simulation environment as expected. Due to the limited on-device computational power and bandwidth, frame loss and runtime will increase inevitably. 

\section{Discussion}
\label{sec:discussion}
\subsection{Open-loop v.s. Closed-loop System}

In this paper, our model is trained and tested in an open-loop manner. Although the results seem acceptable, this manner has certain drawbacks.

First, as Comma.ai has figured out\footnote{\url{https://blog.comma.ai/end-to-end-lateral-planning/}}, the model just learns and predicts the human drivers' most likely behaviour. In datasets collected from real world, the human drivers just keep the vehicle going straight in most of the time, and avoid any possible bad behaviours, such as dangerous driving and traffic offence. In this way, the model will not even have the chance to learn how to recover from mistakes. For example, if we manually feed a video sequence where the vehicle is going over the center line, which of course is a kind of dangerous driving, the model is likely to predict a trajectory that keeps going straight, instead of returning to the correct line. 

Second, we notice that it may introduce a kind of feature leakage. Under certain circumstances, if the vehicle is currently speeding up, the model will predict a longer and faster trajectory, vice versa. Also, if the vehicle is turning left or right, the model will predict a corresponding trajectory that turns left or right. This suggests that the model is learning from the drivers' driving intentions rather than the road characteristics. A typical scenario that reveals this problem is starting-and-stop according to traffic lights. The model may decide to slow down long before it actually sees the red light or the stop line. Then, when the car is slowly starting to move before the red light turns green, the model may decide to speed up quickly. This makes us believe there is a certain kind of temporal feature leakage.

The barrier to closed-loop training and testing is that we simply cannot. Allowing a non-well-trained vehicle driving on the road based on its predicted trajectory is too dangerous and thus not practical. There are two solutions. The first one is introduced in Comma.ai's blog. We can simulate closed-loop training and testing through the WARP mechanism, making the images collected from one perspective look like being collected from a different perspective. The second one is to train and test the model completely in a simulator. However, neither of the solutions are perfect, for the WARP mechanism will introduce image artifacts while the existing simulators~\cite{dosovitskiy2017carla} fail to render realistic images.

\subsection{Imitation Metric}

Though we propose two imitation metrics (\textit{i.e.}, average Euclidean distance error and average precision) in this paper, we believe that the metrics for imitation learning, especially in the field of autonomous driving, remain open. There are two main factors to concern:

First, the ideal trajectory that the model should learn, is not well-defined. Currently we adopt the human drivers' trajectory as the ``ground truth''. It's not a bad choice. However, human drivers may make mistakes. The model may learn these bad examples, which may cause further problems. Moreover, suppose that the leading vehicle is moving slowly. For human drivers, we can choose to overtake it or not, which leads to at least two possible trajectories. Then, how can we decide which trajectory is better and force the model to learn it?

Second, is it necessary to care about the trajectory points that are so far away? Considering that Openpilot inferences at 30 FPS, then theoretically only the trajectory points within 0.033 seconds in the future are important, for the later ones will be overridden by the next prediction. If a vehicle is moving at 120 kph, then it only moves 1.1 meters in 0.033 seconds. Does it mean that we only need to focus on the trajectories within several meters?

\subsection{``Debug'' an End-to-end Model}

It's common knowledge that algorithm designers should pay attention to corner cases and failure cases. Traditionally, in an autonomous driving system where the perception and planning module are apart, we may conclude which part is at fault on failure. However, for an end-to-end model like Supercombo, whose perception and planning module are fused together, how can we debug it on failure? For example, if the model refuses to stop on red light, how can we know if it successfully detects the red light and the stop line?

Frankly speaking, we do not know the answer, either. Of course there are some genetic solutions to make the model robust, like adding more data, making the model deeper and larger, \textit{etc.}, but they do not help solve a specific problem.

\subsection{Reason for Dual-model Deployment Scheme}
In our work, the dual-model deployment scheme is mainly based on two considerations: \textit{On the one hand}, due to the insufficient labeling of the dataset, the model we reimplement only predicts the trajectories, while the original Supercombo also predicts other information including lead, pose, lanes, road, edges, \textit{etc}. These additional predictions are also required by other middleware modules in the Openpilot system. \textit{On the other hand}, as a mass-produced aftermarket product, Openpilot is more than a pure end-to-end neural network. Instead, it's a systematic project, involving software algorithms, communication middleware, and hardware drivers. There is a certain degree of coupling between each module. Therefore, we deploy the model into the existing Openpilot system with the idea of modularity in mind, rather than starting from scratch.

Of course, it is technically possible to discard the dual-model framework. If the reimplemented model can be perfectly aligned with the Supercombo, then we can simply replace it. One possible way to achieve this is to relabel existing datasets and make sure that all the required information is properly labelled.

\section{Conclusions and Future Work}\label{sec:conclusion}
In this report, to discover how Comma.ai manages to achieve L2 assisted driving on a single device, we reimplement the essential Supercombo model from scratch, and test it on public datasets. Experiments suggest that both the original Openpilot and our reimplemented model can perform well on highway scenarios. To test the whole system in real world, we design a dual-model deployment framework. We verify it in CARLA simulation environment, and deploy it on board, proving that our scheme is applicable.
Our work confirms that a simple yet effective L2 assisted driving system can be integrated onto a single board, and it can work well in most cases. Of course, there are still a lot to explore and discuss. We hope this report can give the audience some inspiration.

We believe there are still many open problems that worth exploring in this field. To name a few:
\begin{enumerate}
    \item As end-to-end learning is data-hungry, we may explore new ways to generate and label high-quality data, such as crowdsourcing and auto-labeling.
    
    \item The reason behind the performance gap between the two datasets is not made clear yet.
\end{enumerate}



\bibliography{op_ref}
\bibliographystyle{unsrtnat}


\clearpage
\appendix
\section{Openpilot Analysis}
\label{sec:appendix}

\subsection{Openpilot: A Real-world Test}
In this section, we present the details of how we test Openpilot in real world. These cases are collected in urban, highway, and crosswalks in Shanghai, China. The qualitative and quantitative results, and the typical failure cases are provided below. Full videos can be found in this website \href{https://sites.google.com/view/openpilot-deepdive/home}{\textcolor{blue}{\texttt{https://sites.google.com/view/openpilot-deepdive/home}
}}.

\subsubsection{Qualitative Results} 

\begin{figure}
    \centering
    \begin{subfigure}[c]{0.48\textwidth}
        \centering
        \includegraphics[width=\textwidth]{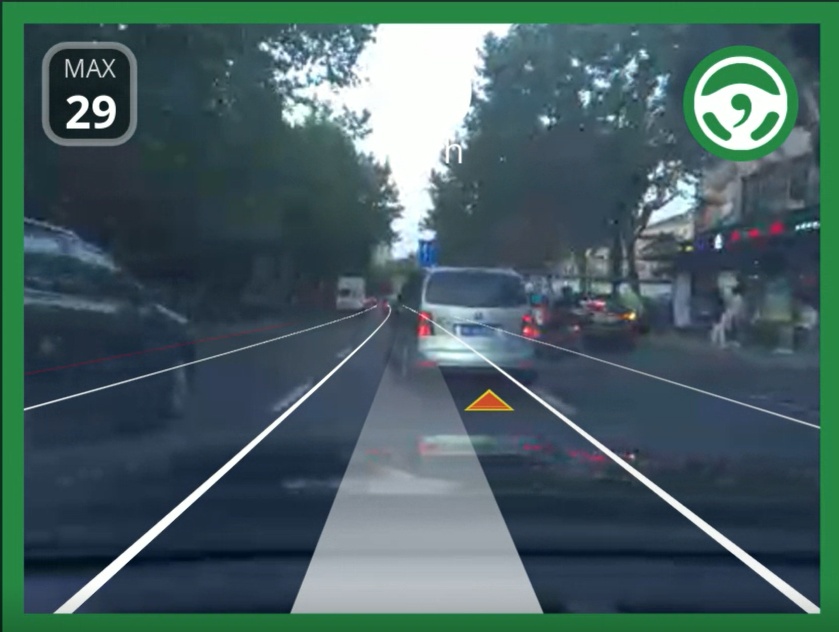}
        \caption{Cut-in.}
        \label{fig:cut_in}
    \end{subfigure}
    \hfill
    \begin{subfigure}[c]{0.48\textwidth}
        \centering
        \includegraphics[width=\textwidth]{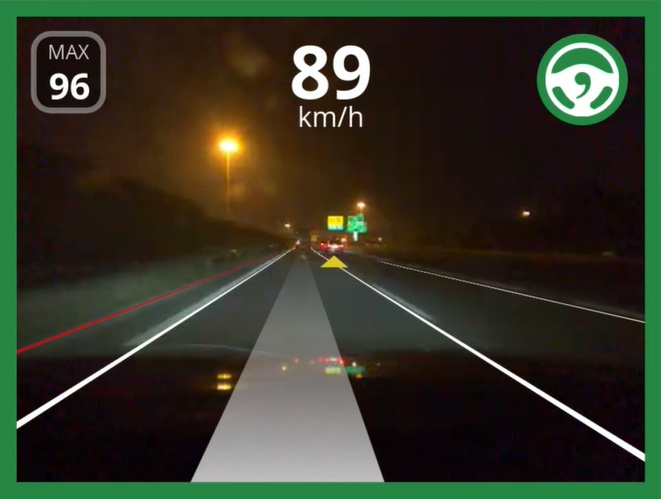}
        \caption{Cut-out.}
        \label{fig:cut_out}
    \end{subfigure}

    \begin{subfigure}[c]{0.48\textwidth}
        \centering
        \includegraphics[width=\textwidth]{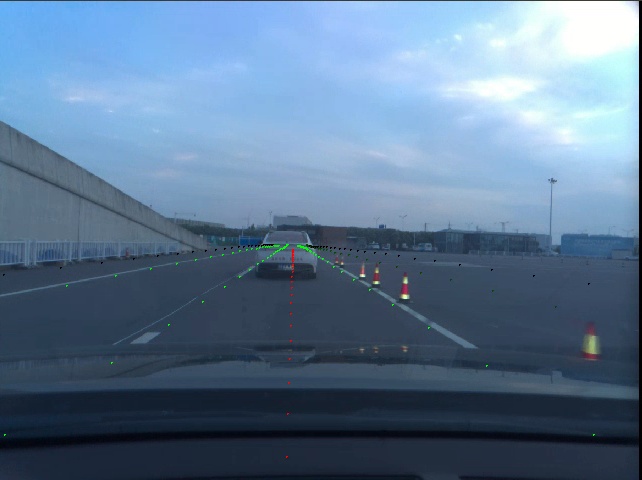}
        \caption{AEB.}
        \label{fig:AEB}
    \end{subfigure}
    \hfill
    \begin{subfigure}[c]{0.48\textwidth}
        \centering
        \includegraphics[width=\textwidth]{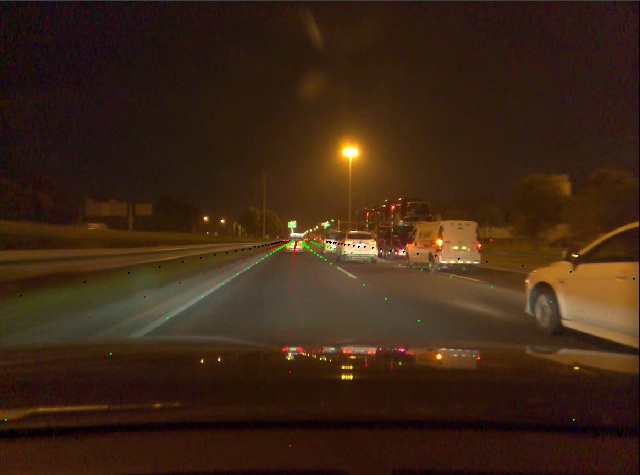}
        \caption{Following-car.}
        \label{fig:following_car}
    \end{subfigure}
    \caption{Tests on ACC.}
\end{figure}

\textbf{ACC.} For the ACC function, we test three different cases, including car cut-in and cut-out, AEB/FCW and car-following. Figure~\ref{fig:cut_in} and Figure~\ref{fig:cut_out} shows typical scenarios of vehicle cut-in and cut-out
Results show that Openpilot can correctly identify the cut-in and cut-out vehicles and change the speed in time when the car is running in a straight road and the speed keeps in the range of 30 and 90 kph.
Also, We test the AEB/FCW function at low speed ($<40$ kph). As shown in Figure \ref{fig:AEB}, when the front car brakes suddenly, Openpilot will slows the car down immediately.
Figure \ref{fig:following_car} shows a typical scenario of a car-following test. We test this function at a full range of speed (from 20 to 120 kph) and under multiple scenarios including straight roads, curves, and intersections. In most cases, Openpilot can successfully follow the leading vehicle.

\begin{figure}
    \centering
    \begin{subfigure}[c]{0.48\textwidth}
        \centering
        \includegraphics[width=\textwidth]{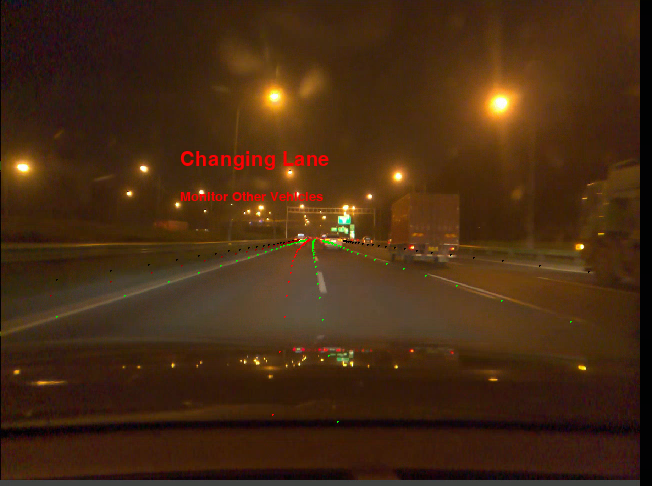}
        \caption{With visible lanelines.}
        \label{fig:ALC_lanelines}
    \end{subfigure}
    \hfill
    \begin{subfigure}[c]{0.48\textwidth}
        \centering
        \includegraphics[width=\textwidth]{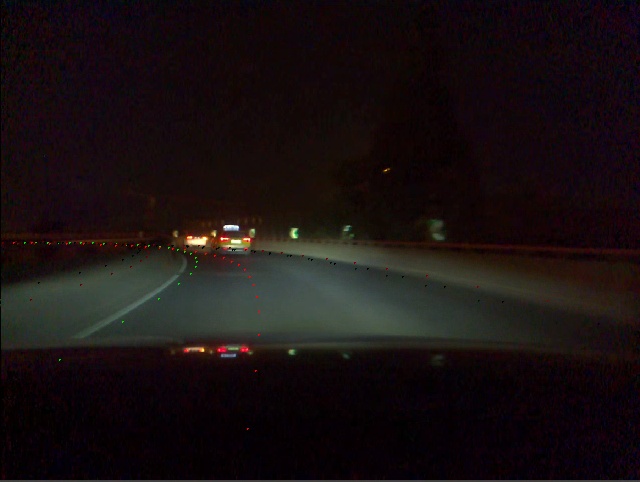}
        \caption{Without visible lanelines.}
        \label{fig:ALC_no_lanelines}
    \end{subfigure}
    \caption{Tests on ALC.}
\end{figure}

\textbf{ALC.} In terms of ALC, we test it in both scenarios where there are visible lane lines and there are no visible lane lines. Results show that when the vehicle speed is maintained above 50 kph, Openpilot can automatically change the driving lane as long as the human driver turns on the turn signal and applies a small force on the steering wheel. However, Openpilot cannot distinguish the types of lane lines, such as dotted lines and solid lines. In other words, it will still try to change the line when there are solid lines. Figure \ref{fig:ALC_lanelines} shows the Openpilot is in an assisted lane change process when there are visible lane lines. Figure \ref{fig:ALC_no_lanelines} shows that Openpilot can still keep the line when there is no visible lane lines in a curved ramp.

\begin{figure}
    \centering
    \begin{subfigure}[c]{0.48\textwidth}
        \centering
        \includegraphics[width=\textwidth]{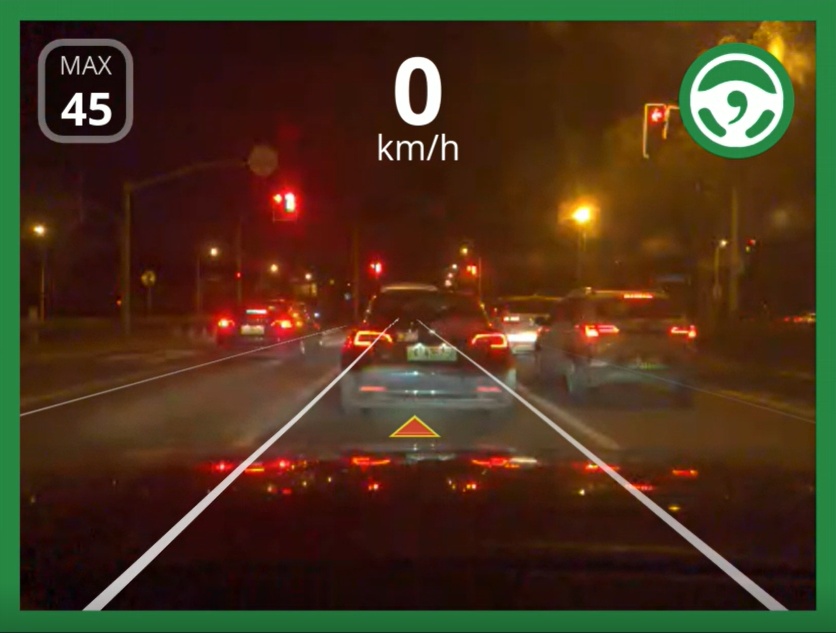}
        \caption{On a straight road with heavy traffic.}
        \label{fig:stop_go}
    \end{subfigure}
    \hfill
    \begin{subfigure}[c]{0.48\textwidth}
        \centering
        \includegraphics[width=\textwidth]{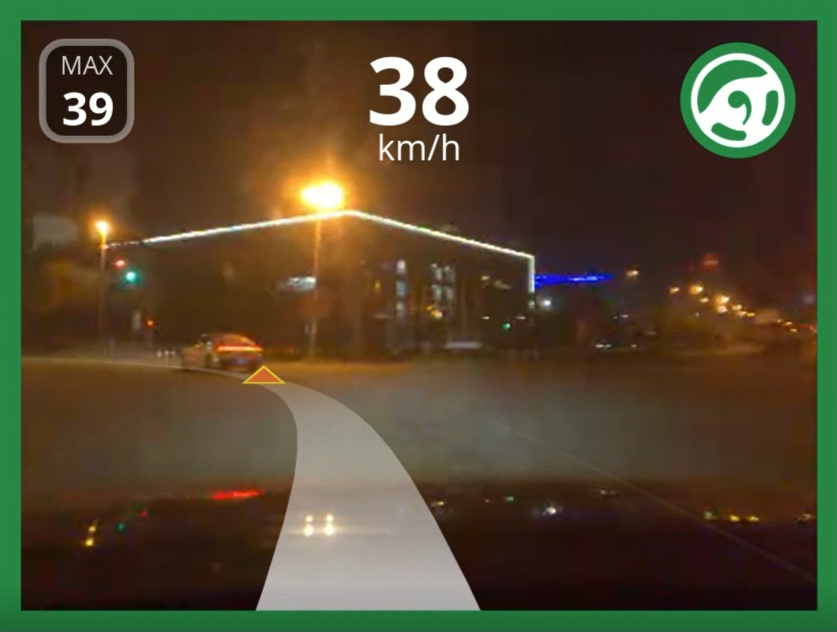}
        \caption{In an intersection turning left.}
        \label{fig:stop_go_turn}
    \end{subfigure}
    \caption{Tests on stop-and-go function.}
\end{figure}

\textbf{Stop-and-Go.} When there is heavy traffic, vehicles may need to start and stop frequently, which may be annoying to human drivers. We test whether Openpilot can well assist human drivers in this case. Figure \ref{fig:stop_go} shows that on a straight road with heavy traffic, Openpilot can tightly follow the leading car when it keeps starting and stopping. The stop-and-go function is also needed when there is an intersection with traffic lights. Figure \ref{fig:stop_go_turn} shows that Openpilot can follow the leading car in an intersection to start and turn left when the red light turns off.

\subsubsection{Quantitative Results}

We also conduct some quantitative experiments to test how accurately Openpilot measures the distance to the driving lane, the distance to the leading car, and the speed of the leading car.

\textbf{Distance to Lane Lines.} In order to measure the error of the distance to lane lines, we stop the car on a straight road, with different distances to the left lane line. Then, we can compare the model predictions with the actual projected distance between the Comma Two device and the line. The results are shown in Table~\ref{tab:dst_lane_line}. We can conclude that Openpilot can accurately predict the position of the lane line.

\textbf{Distance to the Leading Car.} To measure this, we make the vehicle stop on a road. An actor car is placed in different distances to the device. We can then measure the average predicted distances and the standard derivations within a few seconds, and compare with the actual values. The results are reported in Table~\ref{tab:dst_lead}. We can conclude that Openpilot is capable of accurately measuring the distance to the leading car within 50 meters.

\textbf{Speed of the Leading Car.} To measure this, we make the vehicle follow an actor car which is moving at a constant speed. Again, we can measure the averaged speed and the standard derivations within a few seconds, and compare with the actual values. The results are reported in Table~\ref{tab:speed_lead}. We can conclude that Openpilot can accurately measure the speed of the leading car under 60 km/h (16.667 m/s).

\begin{table}[t]
\centering
\caption{Quantitative Results.}
\begin{subtable}{\textwidth}
\centering
\caption{Distance to the Left Lane Line.}
\label{tab:dst_lane_line}
\begin{tabular}{cccccccc}
\toprule
\textbf{Actual Distance} (m) & 0 & -0.12 & -0.32 & -0.49 & -0.78 & -1.02 & -1.20 \\
\midrule
\textbf{Predicted Distance} (m) & -0.05 & 0.10 & -0.30 & -0.56 & -0.79 & -1.14 & -1.32 \\
\bottomrule
\end{tabular}
\end{subtable}

\begin{subtable}{\textwidth}
\centering
\caption{Distance to the Leading Car.}
\label{tab:dst_lead}
\begin{tabular}{ccccccccc}
\toprule
\textbf{Actual Distance} (m/s) & 10 & 20 & 30 & 40 & 50 & 60 & 70 & 80 \\
\midrule
\textbf{Predicted Distance} (m/s) & 10.41 & 19.61 & 29.41 & 34.28 & 51.75 & 56.63 & 58.0 & 63.88 \\
\textbf{std.} (m/s) & 0.27 & 0.58 & 1.04 & 1.81 & 3.21 & 4.09 & 8.07 & 5.12 \\
\bottomrule
\end{tabular}
\end{subtable}

\begin{subtable}{\textwidth}
\centering
\caption{Speed of the Leading Car.}
\label{tab:speed_lead}
\begin{tabular}{cccc}
\toprule
\textbf{Actual Speed} (m/s) & 5.556 & 11.111 & 16.667 \\
\midrule
\textbf{Predicted Speed} (m/s) & 5.45 & 10.13 & 12.86 \\
\textbf{std.} (m/s) & 1.01 & 2.72 & 4.42 \\
\bottomrule
\end{tabular}
\end{subtable}

\end{table}

\subsubsection{Failure Cases}

\begin{figure}
\centering
    \begin{subfigure}[c]{0.48\textwidth}
        \centering
        \includegraphics[width=\linewidth]{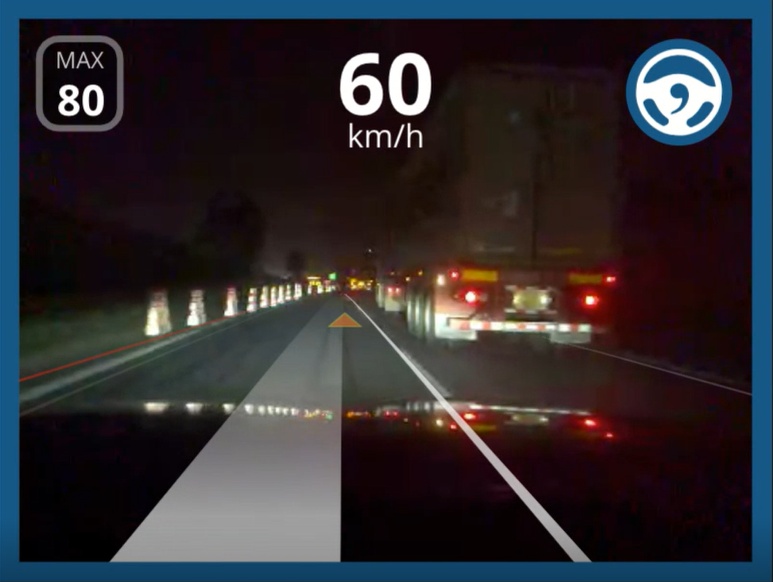}
        \caption{Conical barrels are not detected.}
        \label{fig:failure_case01}
    \end{subfigure}
    \hfill
    \begin{subfigure}[c]{0.48\textwidth}
        \centering
        \includegraphics[width=\linewidth]{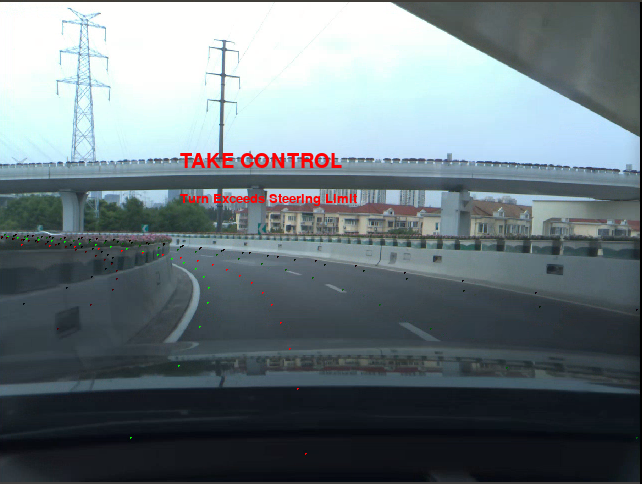}
        \caption{Speed is too fast for the road curvature.}
        \label{fig:failure_case02}
    \end{subfigure}
    
    \begin{subfigure}[c]{0.48\textwidth}
        \centering
        \includegraphics[width=\textwidth]{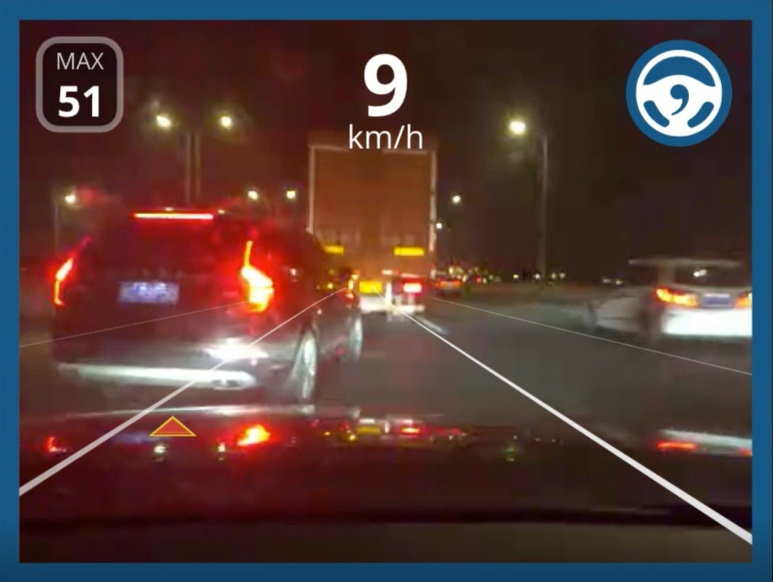}
        \caption{Other car cuts-in closely.}
        \label{fig:failure_case03}
    \end{subfigure}
    \hfill
    \begin{subfigure}[c]{0.48\textwidth}
        \centering
        \includegraphics[width=\textwidth]{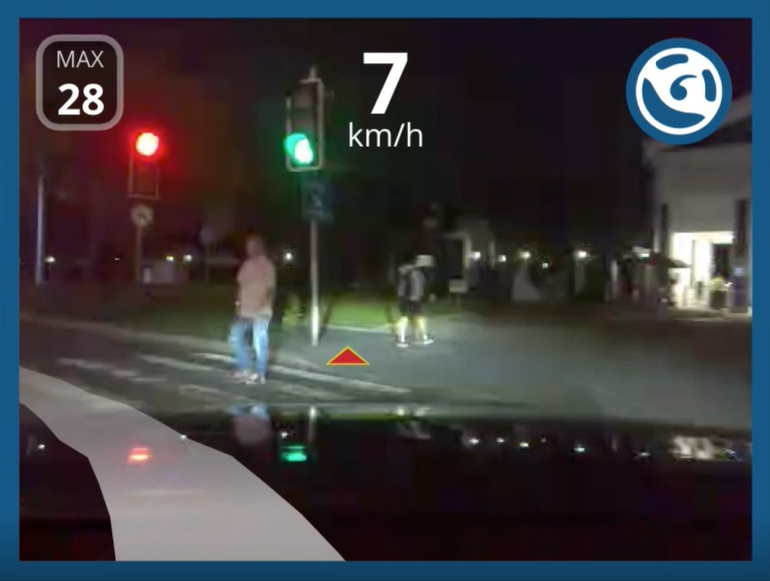}
        \caption{Leading car turns too fast.}
        \label{fig:failure_case04}
    \end{subfigure}

    \begin{subfigure}[c]{0.48\textwidth}
        \centering
        \includegraphics[width=\textwidth]{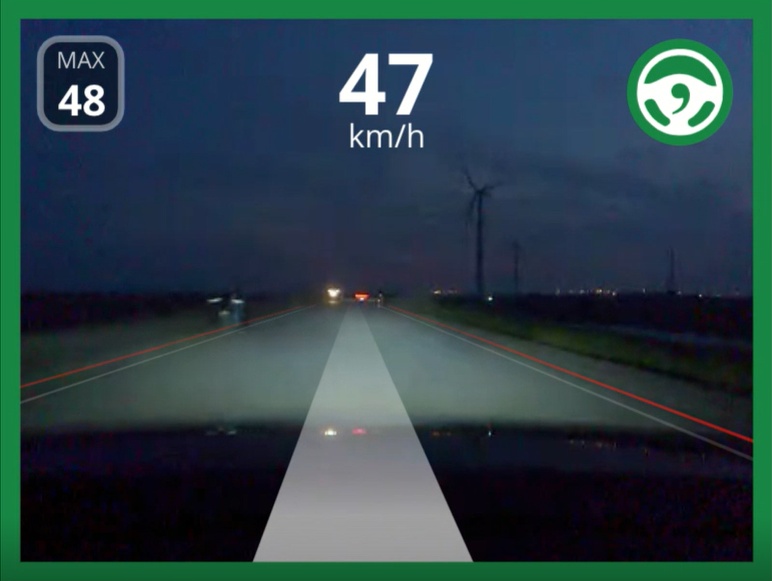}
        \caption{Vehicles are not detected at night.}
        \label{fig:failure_case05}
    \end{subfigure}

\caption{Failure cases.}
\label{fig:failure}
\end{figure}

Although Openpilot shows relatively good performance in most scenarios, still, it may not be good at coping with complex situations. We summarize some typical failure cases. Take Figure \ref{fig:failure_case01} as an example, Openpilot cannot identify objects such as conical barrels, pedestrians, and bicycles on the road. Figure \ref{fig:failure_case02} shows that when the vehicle is fast running through a curved road, Openpilot may not handle it well and may alert the human driver to take control of the vehicle. Figure \ref{fig:failure_case03} shows that even at low speed, when the other car cuts in at a close distance, Openpilot may not respond quickly on time. Figure \ref{fig:failure_case04} shows that Openpilot loses the leading target when following a vehicle to turn, because the leading vehicle quickly vanishes. Figure \ref{fig:failure_case05} shows that Openpilot fails to detect the leading vehicle at night.

\end{document}